%% file: main.tex
\documentclass{clv3}

\usepackage{hyperref}
\usepackage{xcolor}
\definecolor{darkblue}{rgb}{0, 0, 0.5}
\hypersetup{colorlinks=true,citecolor=darkblue, linkcolor=darkblue, urlcolor=darkblue}

\usepackage{CJKutf8}
\usepackage{subcaption}
\usepackage{amsmath}
\usepackage[english]{babel} % auto-wrap

\usepackage{algorithm}
\usepackage[noend]{algpseudocode}
\usepackage{multirow}

\newcommand\seggbmt{SegGBMT}
\newcommand\snrg{GramGBMT}

\DeclareMathOperator*{\argmax}{arg\,max}

\usepackage{tikz}
\usepackage{tikz-qtree}
\usepackage{tikz-dependency}
\usepackage{pgfplots}
\usepackage{pgfkeys}
\usepackage{xparse,xcolor}
\usetikzlibrary{positioning,matrix,arrows,chains,scopes,calc,automata,shapes.geometric,fit,backgrounds}
\usetikzlibrary{patterns}
\usepgfplotslibrary{groupplots}
\pgfplotsset{compat=newest}

\tikzstyle{arrow} = [->,>=stealth]
\tikzstyle{circ} = [draw, circle, inner sep=1.5pt]
\tikzstyle{darrow} = [-implies,double equal sign distance]
\tikzstyle{egnode} = [draw,circle,inner sep=1.5pt]

\pgfkeys {%
    /depgraph/edge unit distance=1.5ex,
}

\let\svtikzpicture\tikzpicture
\def\tikzpicture{\noindent\svtikzpicture}
\runningtitle{Dependency Graph-to-String Translation}

\runningauthor{Li et al.}

\begin{document}

\title{Dependency Graph-to-String Statistical Machine Translation}

\author{Liangyou Li\thanks{Noah’s Ark Lab, Huawei Technologies, Hong Kong. Email: \{liliangyou, qun.liu\}@huawei.com.\newline The work was mainly done when the authors worked in Dublin City University.}}
\affil{Huawei Technologies}

\author{Andy Way\thanks{ADAPT Centre, Dublin City University, Ireland. Email: andy.way@adaptcentre.ie.}}
\affil{Dublin City University}

\author{Qun Liu\footnotemark[1]}
\affil{Huawei Technologies}

\maketitle

%%%%%%%%%%%%%%%%%%%%%%%%%%%%%%%%%%%%%%%%%%

\begin{abstract}

We present graph-based translation models which translate source graphs into target strings. Source graphs are constructed from dependency trees with extra links so that non-syntactic phrases are connected. Inspired by phrase-based models, we first introduce a translation model which segments a graph into a sequence of disjoint subgraphs and generates a translation by combining subgraph translations left-to-right using beam search. However, similar to phrase-based models, this model is weak at phrase reordering. Therefore, we further introduce a model based on a synchronous node replacement grammar which learns recursive translation rules. We provide two implementations of the model with different restrictions so that source graphs can be parsed efficiently. Experiments on Chinese--English and German--English show that our graph-based models are significantly better than corresponding sequence- and tree-based baselines.

\end{abstract}

%%%%%%%%%%%%%%%%%%%%%%%%%%%%%%%%%%%%%%%%%%

\section{Introduction}\label{sec:introduction}

Statistical Machine Translation (SMT) starts from sequence-based models where the basic translation units are words or phrases. IBM made the first breakthrough on SMT by statistically modeling the translation process at the word-level \citep{ibm:brown:1990,ibm:brown:1993}. The well-known phrase-based translation model \citep{pb:koehn:2003} significantly improved upon word-based models by extending translation units from single words to phrases which allow local phenomena, such as word order, word deletion, and word insertion, to be captured. However, conventional phrase-based models are known to be weak at reordering phrases and learning generalizations. For example, assume the following Chinese sentence and its English translation:
\begin{extract}
\begin{CJK*}{UTF8}{gbsn}
\setlength\tabcolsep{4pt}
\begin{tabular}{rlllllll}
     {\it Chinese}: & 2010年 & FIFA & 世界杯 & 在 & 南非 & 成功 & 举行  \\
     {\it Pinyin}: & 2010nian & FIFA & shijiebei & zai & Nanfei & chenggong & juxing \\ 
     {\it Alignment}: & 2010 & FIFA & World Cup & in & South Africa & successfully & held \\
     &&&&&&& \\
     {\it English}: & \multicolumn{7}{l}{2010 FIFA World Cup was held successfully in South Africa} \\
\end{tabular}
\end{CJK*}
\end{extract}
From this example, the phrase-based model learns phrase pairs such as $\langle${\it chenggong juxing, was held successfully}$\rangle$ and $\langle${\it shijiebei, World Cup}$\rangle$. Word reordering inside these phrase pairs is fully captured. However, how to reorder them on the target side to generate a translation is not specified in the model. In addition, generalizations such as $\langle${\it shijiebei \ldots~chenggong juxing, World Cup was held successfully}$\rangle$ are ignored in the phrase-based model because it only uses continuous phrases.

\input{input/fig-exampletree.tex}

Therefore, tree-based (or syntax-based) translation models have been proposed to learn translation rules from tree structures over sentences. For example, given the dependency tree in Figure \ref{fig:tree}, a dependency treelet-based model \citep{treelet:menezes:2005,treelet:quirk:2005,treelet:xiong:2007}, where a treelet is defined as an arbitrarily connected subgraph, can extract the following rule from the aforementioned example:
\begin{align}
\begin{dependency}[hide label,baseline]
    \begin{deptext}[column sep=0cm]
    shijiebei \& chenggong \& juxing \& $\to$ \& World Cup was held successfully\\
    \end{deptext}
    \depedge {3}{1}{}
    \depedge {3}{2}{}
\end{dependency}
\end{align}
where the source side is a tree structure which covers a discontinuous phrase {\it shijiebei \ldots~chenggong juxing}. When a tree-based model is formalized by a synchronous grammar, it will be able to reorder phrases. For example, a dependency tree-to-string model \citep{dep2str:xie:2011} can learn the following translation rule:
\begin{align}
\begin{dependency}[hide label,baseline]
    \begin{deptext}[column sep=0cm]
    $\mathit{NR}_{[1]}$ \& $P_{[2]}$ \& chenggong \& juxing \& $\to$ \& $X_{[1]}$ was held successfully $X_{[2]}$\\
    \end{deptext}
    \depedge {4}{1}{}
    \depedge {4}{2}{}
    \depedge {4}{3}{}
\end{dependency}
\end{align}
where $\mathit{NR}$ and $P$ are source non-terminals representing gaps, $X$ is a general target non-terminal, and indexes indicate mappings between source and target non-terminals and specify how target phrases are reordered when they are inserted into gaps.

However, despite its effectiveness in introducing linguistic knowledge into translation models, syntactic tree structures confine models focusing on linguistically motivated phrases (i.e., syntactic phrases). For example, the dependency treelet-based model only covers phrases which are connected in the tree, while the dependency tree-to-string model only uses phrases which are fully covered by a subtree. Therefore, phrases like {\it 2010nian FIFA} will not be considered in both models. Although linguistically motivated phrases are more reliable in quality and have linguistic meanings, discarding other phrases is a harsh decision which usually does not work well in practice, as these phrases can be quite useful to improve rule coverage and system performance \citep{pb:koehn:2003,nonsyntactic:Hanneman:2009}. 

\input{input/fig-examplegraph.tex}

An obvious observation is that those phrases which are not encouraged in trees
actually are connected in terms of sequential structures (i.e., continuous phrases), such as the phrase {\it 2010nian FIFA}. Since both trees and sequences are special cases of graphs, a possible way of integrating these phrases is using graphs. Therefore, different from previous work which usually incorporates these phrases into tree-based models by using extended labels \citep{spmt:Marcu:2006,ccg:hala:2011,ccg:hala:2012,dep2str:meng:2013,dep2str:aug:xie:2014}, we make a step forward to graph-based models where connected subgraphs are the basic translations units. Figure \ref{fig:graph} shows an example graph which is obtained by adding an edge (from {\it FIFA} to {\it 2010nian}) to the dependency tree in Figure \ref{fig:tree} so that the phrase {\it 2010nian FIFA} will be available in graph-based models. 

In this paper, we explore ways of constructing graphs and design models to translate graphs into strings. Graphs in this paper are constructed by adding edges to dependency trees and thus called {\it dependency graphs}. Dependency trees are used because (i) they directly model syntactic and/or semantic relations between words; (ii) they have the best inter-lingual phrasal cohesion property, i.e., phrases in one language tend to stay together during translation \citep{Fox:2002}; and (iii) we can easily build a large parallel graph--string corpus using dependency parsers. To translate dependency graphs, inspired by phrase-based and tree-based models, we present graph-based models based on graph segmentation and a synchronous grammar. Experiments on Chinese--English (ZH--EN) and German--English (DE--EN) show that our graph-based models are significantly better than their corresponding sequence- and tree-based baselines. 

This paper is based on previous work, including \citet{seggbmt:li:2016} and the PhD thesis of \citet{thesis:liangyou:2017}, by the same authors but with significant differences and contributions. 1) This paper provides formal definitions and more details on graphs and graph-based translation models. 2) This paper introduces a general graph-based model which is based on a synchronous grammar and allows hypotheses covering discontinuous source phrases. Accordingly, a general rule extraction algorithm, a deductive system for decoding and two new features which penalty large distortion and gaps are presented. 3) Because of the exponential complexity when handling graphs during decoding, we present two decoders with different constraints on subgraphs: one is a traditional chart decoder \citep{hpb:chiang:2007} which only considers subgraphs covering continuous source phrases; the other one is a novel beam search decoder which allows subgraphs covering discontinuous source phrases. The beam search-based decoder is our first step towards a general graph-based decoding algorithm. 4) We conducted more experiments and analyze the time complexity of decoding in each experimental system.

In the rest of this paper, we first introduce related work in Section \ref{sec:relatedwork}. Then, we present formal definitions on graphs and introduce two types of dependency graphs (Section \ref{sec:graph}). After that, we describe a segmentation-based model which segments a graph into a sequence of disjoint subgraphs and generates translations by combining subgraph translations (Section \ref{sec:seggbmt}). In Section \ref{sec:snrg}, we further introduce a model based on a synchronous graph grammar which enables our model to learn recursive translation rules. Our experimental results are demonstrated in Section \ref{sec:exp}. Finally, Section \ref{sec:conclusion} summarizes our work and discusses possible avenues for future research.

%%%%%%%%%%%%%%%%%%%%%%%%%%%%%%%%%%%%%%%%%%

\section{Related Work}\label{sec:relatedwork}

According to the fundamental structures used, we divide different translation models into three categories: sequence-based models, tree-based models and graph-based models.

\subsection{Sequence-Based Models}

Since the breakthrough made by IBM on word-based models in the 1990s \citep{ibm:brown:1990,ibm:brown:1993}, SMT has developed rapidly. The phrase-based model \citep{pb:koehn:2003} advanced the state-of-the-art by translating multi-word units, which makes it better able to capture local phenomena within phrases. However, it cannot reorder the phrases themselves. Even though reordering models \citep{pb:koehn:2003,wmsd:koehn:2005,maxent:Xiong:2006,hrlm:galley:2008,sparse:cherry:2013} can be used to guide the phrase reordering, it is still known to be weak at long-distance reordering. Another disadvantage is that only continuous phrases are considered, and thus the learned translation pairs cannot be generalized even though sometimes an apparent pattern can be recognized. \citet{DTU:Galley:2010} extended the phrase-based model by allowing phrases with gaps (i.e., discontinuous phrases). However, without using linguistic knowledge, the model can learn plenty of unreliable translation rules.

\subsection{Tree-Based Models}

Tree-based models are proposed to alleviate problems in sequence-based models by learning translation rules which allow phrase reordering and generalization.

\subsubsection{Hierarchical Phrase-Based Models}

A hierarchical phrase is an extension of a phrase by allowing gaps where other hierarchical phrases are embedded \citep{hpb:chiang:2005,hpb:chiang:2007}. The hierarchical phrase-based (HPB) model \citep{hpb:chiang:2005,hpb:chiang:2007} is formulated by a {\it synchronous context-free grammar} (SCFG) with only one general non-terminal $X$. Even though it provides the model with more flexibility, the only non-terminal $X$ often makes it hard for the model to select the most appropriate rules. Therefore, some work refines this non-terminal using linguistic information, such as syntactic categories from constituent structures \citep{syntaxhpb:Zollmann:2006}, POS tags or word classes \citep{classhpb:Zollmann:2011}, supertags based on combinatory categorical grammars (CCGs) \citep{ccg:hala:2011,ccg:hala:2012}, and head information from dependency structures \citep{head:junhui:2012}.

\subsubsection{Constituent Tree-Based Models}

A constituent (or phrasal) structure displays the functional components of a sentence. Typically, models based on constituent trees are formulated in a {\it synchronous tree-substitution grammar} (STSG) \citep{stsg:Eisner:2003}. \citet{ghkm:galley:2004,ghkm:galley:2006} proposed a well-known string-to-tree model based on STSG which translates source sentences into target trees. Given word-aligned string-tree pairs, this model automatically extracts transfer rules which map source phrases into target tree fragments. Similarly but differently, tree-to-string models \citep{t2s:huang:2006,t2s:Huang:2006:amta,t2s:liu:2006} use parse trees on the source side. Compared with string-to-tree models, tree-to-string models can decode a sentence in linear time in practice with respect to the sentence length \citep{t2s:Huang:2010}. Tree-to-tree models \citep{t2t:MinZhang:2007,t2t:Nesson:2006} use trees on both sides. Despite benefits brought by linguistic trees, constituent tree-based models have severe problems on integrating non-syntactic phrases which are not linguistically well-formed but can be important to translation performance of systems \citep{pb:koehn:2003,nonsyntactic:Hanneman:2009,nonsyntactic:matthias:2014}. To make use of such phrases, additional non-terminal symbols \citep{spmt:Marcu:2006,syntaxhpb:Zollmann:2006,ccg:hala:2011,ccg:hala:2012} or binarization of syntax trees \citep{binarize:zhang:2006,binarize:wang:2007} may be needed. However, such kinds of relaxation of syntactic constraints can result in less grammatical translations \citep{syscomp:Kaljahi:2012}.

\subsubsection{Dependency Tree-Based Models}

Dependency structures directly model relations between words in a sentence, each of which indicates the syntactic and/or semantic function of one word in relation to another word. A dependency tree can be segmented into a set of elementary units, such as edges, paths, or treelets which can be used in SMT. \citet{path:Lin:2004} proposed a dependency path-based model which translates a source dependency tree by combining translations of each path. The treelet approach \citep{treelet:menezes:2005,treelet:quirk:2005} translates a dependency tree by bottom-up combining translations of disjoint treelets. \citet{treelet:xiong:2007} extended the treelet approach to allow gaps. \citet{deppath:chen:2014} proposed an edge-based model where the basic translation units are dependency edges. However, in these models, translation rules do not encode enough reordering information. By contrast, models based on synchronous grammars have proven to be better capable of handling phrase reordering. \citet{str2dep:shen:2010} presented a model which is based on the HPB model with an extension using target dependency trees. Because the model only considers structures which cover continuous spans, it is easier to integrate a dependency-based language model which can significantly improve the system. Different from string-to-dependency models, \citet{dep2str:xie:2011} presented a dependency tree-to-string model with an extended SCFG by including dependency links. However, the model only considers syntactic phrases. Since dependency trees are flatter than constituent trees, this model has a severe data-sparsity problem \citep{dep2str:meng:2013,dep2str:aug:xie:2014}. To incorporate non-syntactic phrases, \citet{dep2str:meng:2013} proposed to simultaneously use dependency trees and constituent trees so that phrases which are non-syntactic in dependency trees but syntactic in constituent trees can be covered. \citet{dep2str:aug:xie:2014} incorporated fixed and floating structures into the dependency tree-to-string model by creating special labels at run-time. \citet{dep2str:liangyou:2014} extended this model by decomposing dependency structures so that the translation of a syntactic phrase can be generated by combining translations of subphrases inside it.

\subsection{Graph-Based Models}

Graphs are more general and powerful representations than trees and thus believed to be better able to capture sentence meanings. In recent year, {\it abstract meaning representation} (AMR) \citep{amr:banarescu:2013} has been widely investigated which uses hypergraphs to represent semantic meanings of sentences. \citet{semanticmt:jones:2012} presented a semantics-based translation model, where a source sentence is firstly parsed into a hypergraph using a {\it synchronous hyperedge replacement grammar} (SHERG) and then the hypergraph is transformed into a target string using a target SHERG. However, the recognition algorithm for SHERG is in polynomial time but potentially of a high degree \citep{hypergraph:Lautemann:1990,hypergraph:chiang:2013}. Furthermore, large parallel corpora annotated with hypergraphs are not readily available. 

Compared to tree-based models which are usually based on a binary SCFG such as the HPB model and allow only phrasal discontinuities, graph-based models use subgraphs as basic translation units which may cover discontinuous phrases and thus have more powerful expressiveness than tree-based models \citep{DTU:Galley:2010}.

%%%%%%%%%%%%%%%%%%%%%%%%%%%%%%%%%%%%%%%%%%

\section{Dependency Graphs}\label{sec:graph}

Graphs used in this paper are called {\em dependency graphs} which are node-labeled, directed and connected. We do not consider edge labels as they did not improve translation performance in our experiments (Section \ref{sec:exp}) and will complicate our explanation. Before introducing how dependency graphs are constructed, we first provide a formal definition:
\begin{definition}\label{def:graph}
A node-labeled and directed {\it dependency graph} (or graph for short) is a tuple $\langle V,E,\phi\rangle$, where $V$ is a finite set of nodes, $E\subseteq V^2$ is a finite set of edges, and $\phi: V\to C$ is a function which assigns a label from $C$ to each node.
\end{definition}

\input{input/fig-nodeorder.tex}

For simplicity, from now on we use the terms graph and dependency graph interchangeably. Note that although in Definition \ref{def:graph} nodes are unordered, we will assume that nodes are ordered according to word order as this is an important source of information for SMT. Figure \ref{fig:nodeorder} shows two example graphs which have different node order and thus are different. The basic translation units in our graph-based models are {\em node-induced subgraphs}, which are connected and defined as follows.

\begin{definition}\label{def:subgraph}
A {\it node-induced subgraph} of a graph $\langle V,E,\phi\rangle$ is a graph $\langle V',E',\phi'\rangle$, where $V'\subseteq V$, $E'\subseteq E$, $\forall u\in V':\phi'(u)=\phi(u)$, and $\forall u,v\in V':(u,v)\in E\Leftrightarrow (u,v)\in E'$.
\end{definition}

\input{input/fig-subgraph.tex}

According to Definition \ref{def:subgraph}, a node-induced subgraph is a subset of nodes of a graph together with all edges whose endpoints are both in this subset. Figure \ref{fig:subgraph} provides examples of two subgraphs of Figure \ref{fig:nodeorder:a} with one of them node-induced. Since in this paper we only deal with node-induced subgraphs, from now on, we will assume that all subgraphs are node-induced subgraphs and use the two terms without distinction.

% \begin{theorem}\label{def:subgraph:node}
% Given a graph $G=\langle V,E,\phi\rangle$ and a subset $V'\subseteq V$ of nodes, the node-induced subgraph $\langle V',E',\phi'\rangle$ of $G$ is unique.
% \end{theorem}
% \begin{proof}
% Assuming there are two different node-induced subgraphs: $\langle V',E',\phi'\rangle$ and $\langle V',E'',\phi'\rangle$. Then, one of them must have more edges.  $\exists e\in E'$
% \end{proof}

\subsection{Dependency-Bigram Graphs}\label{sec:graph:dbg}

\input{input/fig-dbg.tex}

The first kind of graph used in this paper directly combines a sequence and a dependency tree by using bigram links and dependency links. The graph is therefore called a {\em dependency-bigram graph} (DBG). Figure \ref{fig:dbg} shows an example DBG. Each edge in the DBG denotes either a bigram relation or a dependency relation. Bigram relations are implied in sequences and provide local and sequential information on pairs of continuous words. Phrases connected by bigram relations, i.e., continuous phrases, are known to be useful for improving phrase coverage \citep{nonsyntactic:Hanneman:2009}. By contrast, dependency relations come from dependency structures which model syntactic and/or semantic relations between words. Phrases connected by dependency relations are covered by treelets and thus more linguistically motivated and reliable \citep{treelet:quirk:2005}. By combining the two kinds of relations together, we can make use of both continuous and linguistically-informed discontinuous phrases without distinction as long as they are covered by subgraphs.

\input{input/fig-dbgphrtype.tex}

For instance, given the graph in Figure \ref{fig:dbg}, we can use subgraphs which are connected by different combinations of links as in Figure \ref{fig:dbgphrtype}. These subgraphs cover three kinds of phrases:
\begin{enumerate}
\item Phrases as in Figures \ref{fig:dbgphrtype:a}--\ref{fig:dbgphrtype:c} which are connected in terms of bigram links. These phrases are continuous and also used in phrase-based models.
\item Phrases as in Figures \ref{fig:dbgphrtype:b}--\ref{fig:dbgphrtype:d} which are connected in terms of dependency links. These phrases can also be used in dependency treelet-based models.
\item Phrases as in Figure \ref{fig:dbgphrtype:e} which are only connected when both types of links are considered. These phrases cannot be covered by both phrase-based systems and dependency treelet-based systems.
\end{enumerate}
In experiments, we found $\sim$70\% of rules are extracted from continuous phrases on both ZH--EN and DE--EN in our segmentation-based model (Section \ref{sec:seggbmt}). This also means that source sides of most rules are connected by bigram links. Around 42\%–48\% of rules are connected by dependency links. We also observed that >30\% of rules are connected by not only bigram links but also dependency links. About 15\%–17\% of rules falls into the third category which slightly improve our model resulting in the best translation performance.

\subsection{Dependency-Sibling Graphs}\label{sec:graph:dsg}

Another kind of graph used in this paper is called a {\em dependency-sibling graph} (DSG) which is constructed by adding sibling links to a dependency tree. Figure \ref{fig:dsg} shows an example DSG. Compared with bigram relations used in DBGs, phrases connected by sibling relations are usually fewer in number but more linguistically motivated. Given the graph in Figure \ref{fig:dsg}, we can also use subgraphs covering three kinds of phrases: (i) phrases which are connected by sibling links and thus may be discontinuous as in Figure \ref{fig:dsgphrtype:a}; (ii) phrases as in Figure \ref{fig:dsgphrtype:b} which are connected by dependency links; (iii) phrases as in Figure \ref{fig:dsgphrtype:c} which are not connected by a single type of link.

\input{input/fig-dsg.tex}

\input{input/fig-dsgphrtype.tex}

By comparing Figure \ref{fig:dbgphrtype} and Figure \ref{fig:dsgphrtype}, we can see that some phrases covered by subgraphs are shared by the DBG and DSG (e.g., Figure \ref{fig:dbgphrtype:d} and Figure \ref{fig:dsgphrtype:b}). However, there are also some phrases which are only available in either a DBG (e.g., {\it Nanfei chenggong} as in Figure \ref{fig:dbgphrtype:a}) or a DSG (e.g. {\it zai chenggong} as in Figure \ref{fig:dsgphrtype:a}). This is because while a DBG includes links between any two consecutive words, a DSG only adds links to two consecutive siblings which may be discontinuous in a large sequence. In addition, even though some phrases are available in both a DBG and DSG, they have different graph structures, e.g., Figure \ref{fig:dbgphrtype:e} and Figure \ref{fig:dsgphrtype:c}.

%%%%%%%%%%%%%%%%%%%%%%%%%%%%%%%%%%%%%%%%%%

\section{Graph Segmentation-Based Translation}\label{sec:seggbmt}

In this section, we present a segmentational graph-based translation model (called {\it\seggbmt}). Inspired by phrase-based models, our model segments an input graph into a sequence of disjoint subgraphs and generates a complete translation by combining translations of each subgraph left-to-right using beam search. In the following subsections, we firstly introduce notation and some definitions (Section \ref{sec:seggbmt:def}) which will also be used in Section \ref{sec:snrg}. Then, we present a training algorithm (Section \ref{sec:seggbmt:train}), features (Section \ref{sec:seggbmt:model}) and a decoding process (Section \ref{sec:seggbmt:decode}).

\subsection{Notation and Definitions}\label{sec:seggbmt:def}

Let $\langle G(s),t,a\rangle$ be a parallel graph--string pair, where $G$ is a graph covering a source sentence $s$, $t$ is a target sentence, and $a$ is a set of mappings between positions of $s$ and positions of $t$. We use $s_i$ and $t_j$ to denote individual words at a source position $i$ and a target position $j$, respectively. We denote a source discontinuous phrase as $\tilde{s}=\bar{s}_1\bar{s}_2\cdots\bar{s}_K$ which contains $K$ continuous phrases $\bar{s}_1\cdots\bar{s}_K$ and thus $K-1$ gaps. When $K=1$, $\tilde{s}$ is a continuous phrase. A subraph covering $\tilde{s}$ is denoted as $G(\tilde{s})$ if it exists. We use $\bar{t}$ to represent a target continuous phrase. To simplify the terminology used to explain our models, from now on, the term {\em phrase} only represents continuous phrases while a {\em subsequence} can be either a continuous phrase or a discontinuous phrase.

Our \seggbmt~model can be seen as an extension of the phrase-based model by taking subgraphs as the basic translation units, as in Equation (\ref{eq:seggbmt}):
\begin{equation}\label{eq:seggbmt}
\begin{split}
    p(G(\tilde{s}_1^I) \mid \overline{t}_1^I) &= \prod_{i=1}^{I} p(G(\tilde{s}_{a_i}) \mid \overline{t}_i) 
    d(G(\tilde{s}_{a_i}), G(\tilde{s}_{a_{i-1}}))\\
    &\approx \prod_{i=1}^{I} p(G(\tilde{s}_{a_i}) \mid \overline{t}_i) 
    d(\tilde{s}_{a_i}, \tilde{s}_{a_{i-1}})
\end{split}
\end{equation}
where $d(\cdot)$ is a distortion function as in the phrase-based model which will be defined in Section \ref{sec:seggbmt:model}. According to Equation (\ref{eq:seggbmt}), a target sentence is segmented into a sequence of $I$ phrases in \seggbmt. Each $\bar{t}_i$ is a translation of a source subgraph $G(\tilde{s}_{a_i})$. Accordingly, the sequence of subgraphs $[G(\tilde{s}_{a_1}),\cdots,G(\tilde{s}_{a_I})]$ is called a {\it graph segmentation}, where subgraphs are disjoint with each other, as in Definition \ref{def:graphseg}. 

\begin{definition}\label{def:graphseg}
A {\it segmentation} of a graph $\langle V,E,\phi\rangle$ is a sequence of disjoint subgraphs $[\langle V_1,E_1,\phi_1\rangle,\cdots,\langle V_I, E_I, \phi_I\rangle]$, where $V_1\cup\cdots\cup V_I=V$.
\end{definition}

Note that during segmention of a graph, nodes are divided into subgraphs and edges between subgraphs are ignored. Therefore, subgraphs in a graph segmentation cover all nodes rather than edges. This also means that when subgraphs in a graph segmentation are combined to form a graph, their nodes remain disjoint and new edges are formed between them. Figure \ref{fig:dbg2str} shows a graph--string pair where the graph is segmented into three subgraphs, each of which corresponds to a target phrase.

\input{input/fig-dbg2str.tex}

\subsection{Rule Extraction}\label{sec:seggbmt:train}

\input{input/fig-subgraphphrase.tex}

Different from phrase-based models, the basic translation units in our \seggbmt~model are subgraphs. Accordingly, given a parallel graph--string pair $\langle G(s),t,a\rangle$, we extract {\it subgraph--phrase pairs} $\langle G(\tilde{s}),\overline{t}\rangle$ as translation rules, which are consistent with the word alignment $a$ \citep{align:consistent:och:2004}. An example subgraph-phrase pair extracted from the running example using the DBG in Figure \ref{fig:dbg} is shown in Figure \ref{fig:subgraphphrase}. It translates a source subgraph into a target phrase {\it 2010 FIFA World Cup}. We now provide a formal definition of a {\it subgraph--phrase pair}.
\begin{definition}\label{def:spp}
Given a graph--string pair $\langle G(s),t,a\rangle$, let $\bar{t}$ be a phrase of $t$ and $G(\tilde{s})$ be a subgraph of $G(s)$ covering a source subsequence $\tilde{s}$, $\langle G(\tilde{s}),\overline{t}\rangle$ is a subgraph--phrase pair of $\langle G(s),t,a\rangle$, iff $\langle\tilde{s},\bar{t}\rangle$ is consistent with $a$, i.e.:
\begin{enumerate}
\item $\exists s_i\in\tilde{s}$, $t_{j}\in\bar{t}$: $(i,j)\in a$.
\item $\forall s_i\in\tilde{s}$: $(i,j)\in a \Rightarrow t_{j}\in\bar{t}$.
\item $\forall t_{j}\in\bar{t}$: $(i,j)\in a \Rightarrow s_i\in\tilde{s}$.
\end{enumerate}
\end{definition}

\input{input/fig-sppair.tex}

Note that the source side of a subgraph-phrase pair in \seggbmt~is a subgraph which is connected and does not contain any non-terminals. The subgraph can be used to cover either a continuous phrase or a discontinuous phrase. The target side of the pair is always a phrase. Figure \ref{fig:invalidsppair} shows two pairs which are not considered as subgraph--phrase pairs: in Figure \ref{fig:invalidsppair:a} the pair is not consistent with the word alignment, and in Figure \ref{fig:invalidsppair:b} the source side is not a subgraph of the DBG in Figure \ref{fig:dbg}.

The procedure of extracting subgraph-phrase pairs is as follows:
\begin{algorithmtext}
    \item [Step 1] Find a new target phrase $\bar{t}:|\bar{t}|\le L$;
    \item [Step 2] Find all source subsequences $Q=\{\tilde{s}\mid \forall t_j\in\bar{t}: (i,j)\in a\Rightarrow s_i\in\tilde{s} \text{ and } |\tilde{s}|\le L\}$;
    \item [Step 3] Pop an element $\tilde{s}$ from $Q$;
    \item [Step 4] If $\langle\tilde{s},\bar{t}\rangle$ is consistent with $a$ and $G(\tilde{s})$ exists, $\langle G(\tilde{s}),\bar{t}\rangle$ is a subgraph-phrase pair;
    \item [Step 5] Go back to Step 3 until $Q$ is empty;
    \item [Step 6] Go back to Step 1 until all target phrases have been visited.
\end{algorithmtext}
This procedure traverses each pair $\langle\tilde{s},\overline{t}\rangle$, which is within a length limit $L$ ($L=7$ in our experiments) and consistent with the word alignment $a$, and outputs $\langle G(\tilde{s}),\overline{t}\rangle$ if $\tilde{s}$ is covered by a subgraph $G(\tilde{s})$. A source subsequence can be extended with unaligned source words which are adjacent to it on boundaries so that all phrases which are consistently aligned to the same target phrase can be accessed.

\subsection{Model and Features}\label{sec:seggbmt:model}

We define our model in the log-linear framework \citep{loglinear:och:2002} over a derivation $d=r_1r_2\cdots r_N$, as in Equation (\ref{eq:loglinear}):
\begin{equation}
\label{eq:loglinear}
p(d) \propto \prod_i \phi_i (d)^{\lambda_i}
\end{equation}
where $r_i$ are translation rules, $\phi_i$ are features defined on derivations, and $\lambda_i$ are feature weights. In our experiments, we use the following standard features: 
\begin{itemize}
    \item Two translation probabilities $p(G(s)|t)$ and $p(t|G(s))$ based on frequency;
    \item Two lexical translation probabilities $p_{\mathrm{lex}}(s|t)$ and $p_{\mathrm{lex}}(t|s)$ based on word alignment \citep{pb:Och:1999};
    \item A language model $p(t)$ to score a translation $t$;
    \item A rule penalty $\exp(-N)$;
    \item A word penalty $\exp(-|t|)$;
    \item A distortion penalty $\exp(-d(\cdot))$ for distance-based reordering.
\end{itemize}

The calculation of the distortion function $d(\cdot)$ in our model is different from the one in conventional phrase-based models, because we need to take discontinuity into consideration. In our model, we use a distortion function as in Equation (\ref{eq:seggbmt:distortion}) to penalize discontinuous phrases that have relatively long gaps \citep{DTU:Galley:2010}: 
\begin{equation}
\label{eq:seggbmt:distortion}
\begin{split}
d(\tilde{s}_{a_i}, \tilde{s}_{a_{i-1}}) = |\tilde{s}_{a_i}^{b} - \tilde{s}_{a_{i-1}}^{e} -1| + \sum_{k=2}^K |\overline{s}_{a_i,k}^{b} - \overline{s}_{a_i,k-1}^{e} -1|
\end{split}
\end{equation}
where superscripts $b$ and $e$ denote the beginning and end positions of a subsequence, respectively. $\tilde{s}_{a_i}=\bar{s}_{a_i,1}\cdots\bar{s}_{a_i,K}$ is a subsequence which has $K-1$ gaps and thus consists of $K$ phrases $\overline{s}_{a_i,k}$. Figure \ref{fig:seggbmt:distortion} shows an example of calculating distortion values. According to Equation (\ref{eq:seggbmt:distortion}), the value of the distortion function $d$ is a summation (denoted by $+$) of two values. While the first value measures the distance between the current and a previous subsequence, the second value calculates the length of gaps in the current subsequence. In practice, instead of adding them together as a single distortion value, we treat the two values as two distinct features \citep{DTU:Galley:2010} so in experiments there are in total 9 features in our \seggbmt~model. 

\input{input/fig-seggbmt-distortion.tex}

\subsection{Decoding}\label{sec:seggbmt:decode}

During decoding, the \seggbmt~model searches for the best derivation $\hat{d}$ whose source yield $f(\hat{d})$ corresponds to a segmentation of $G(s)$ denoted by a $\asymp$ relation and target yield $e(\hat{d})$ is a target sentence $\hat{t}$, as in Equation (\ref{eq:seggbmt:decode}):
\begin{equation}\label{eq:seggbmt:decode}
\hat{t} = e\left(\argmax_{d\in D : f(d)\asymp G(s)} p(d)\right)
\end{equation}

\input{input/fig-seggbmt-decode.tex}

The decoder in \seggbmt~is similar to the phrase-based decoder, which generates hypotheses (partial translations) from left to right using beam search. Each hypothesis maintains a {\em coverage vector} and can be extended by translating an uncovered subgraph. Positions covered by the subgraph are then marked as translated. The translation process ends when no untranslated words remain. Hypotheses in the same stack can be recombined and pruned according to their partial translation cost and an estimated future cost \citep{pb:koehn:2003,DTU:Galley:2010}. Figure \ref{fig:seggbmt:decode} shows a derivation of translating an input DBG in Chinese to an English string.

\input{input/alg-seggbmtdecoder.tex}

The decoding procedure for \seggbmt~is shown in Algorithm \ref{alg:seggbmt:decode}. The algorithm maintains $|s|+1$ stacks $B_0,B_1,\cdots ,B_{|s|}$. Each $B_i$ contains a set of hypotheses covering exactly $i$ source words. The algorithm starts from an empty hypothesis $h_\emptyset$ which does not cover any words (Line 1). Then, it traverses each stack $B_i$ and each hypothesis $h_c$ in $B_i$ (Lines 2--3) where the coverage vector $c$ maintains the set of positions already covered by the hypothesis. To extend $h_c$, the algorithm considers all translation rules which are within a {\em distortion limit} of $d_{\text{max}}$ (6 in our experiments) in terms of the first uncovered position and do not overlap with already covered source words (Lines 4--5). Given an applicable rule $r=\langle G(\tilde{s}),\bar{t}\rangle$, the function $\mathrm{Create}$ extends $h_c$ to generate a new hypothesis $h_{c'}$ by appending $\bar{t}$ to the right and updating the coverage vector and weights (Line 6). Then new hypothesis is then added to a stack according to the number of words covered (Line 7). When all stacks have been visited, the decoder returns the best hypothesis in stack $B_{|s|}$ as the final translation.

% {\bf TODO: time complexity?}

% Although time complexity of the phrase-based decoder with beam search is in practice $O(|s|bd_{\text{max}})$ with a bounded phrase size \citep{beamt2s:huang:2010}, our decoder is more time-consuming as we translate subgraphs which cover subsequences rather than phrases. In theory, the number of subgraphs within a bounded size $L$ of a graph is exponential to the sentence length. XXXX

%%%%%%%%%%%%%%%%%%%%%%%%%%%%%%%%%%%%%%%%%%

\section{Synchronous Grammar-Based Translation}\label{sec:snrg}

In Section \ref{sec:seggbmt}, we presented a graph-based translation model which only uses non-recursive rules and generates a translation by segmenting a graph and combining subgraph translations. Although the model naturally takes both continuous phrases and discontinuous phrases into consideration, it is difficult to reorder target phrases. Therefore, in this section, we introduce a new model (called {\it\snrg}) which uses a synchronous graph grammar to parse input graphs and simultaneously generate target strings. Translation rules in \snrg~may contain non-terminals which are used to specify how target phrases are reordered.

In the following sections, we firstly introduce the grammar (Section \ref{sec:snrg:grammar}). Then, we present an algorithm to extract translation rules (Section \ref{sec:snrg:train}), features (Section \ref{sec:snrg:model}) and a deductive proof system \citep{proofsys:shieber:1995,proofsys:goodman:1999} for decoding with two different implementations (Section \ref{sec:snrg:decode}).

\subsection{Grammar}\label{sec:snrg:grammar}

Similar to tree grammars which are used to generate trees, graph grammars are rewriting formalisms for generating graphs. In this section, we present a {\em synchronous node replacement grammar} (SNRG) which generates graph pairs by replacing nodes with other graphs. It will be used in this paper to translate graphs. In an SNRG, the elementary units are graph fragments, which are also the right-hand sides of production rules in the grammar. Its definition is as follows:
\begin{definition}
\label{def:graphfrag}
\normalfont A {\em graph fragment} is a tuple $\langle V,E,\phi,\eta\rangle$, where $\langle V,E,\phi\rangle$ is a graph and $\eta$ is an embedding mechanism which consists of a set of connection instructions to indicate how to add edges when integrating a graph into another graph.
\end{definition}

\input{input/fig-graphfrag.tex}

Figure \ref{fig:graphfrag} shows an example graph fragment which consists of a graph (on the left) and an embedding mechanism (on the right). According to the embedding mechanism, when we integrate the graph into another one, if {\it juxing} and {\it zai} are neighbours of the graph, two edges ({\it juxing $\to$ shijiebei}) and ({\it zai $\to$ shijiebei}) will be added. Based on graph fragments, we define an SNRG as in Definition \ref{def:snrg}.

% \begin{definition}
% \label{def:nrg}
% \normalfont A {\em node replacement grammar} (NRG) is a tuple $\langle N,T,P,S\rangle$, where $N$ and $T$ are disjoint finite sets of non-terminal symbols and terminal symbols, $S\in N$ is the start symbol, and $P$ is a finite set of productions of the form $\left(A\rightarrow R\right)$. $A\in N$, and $R$ is graph fragment where node labels are from $N\bigcup T$.
% \end{definition}

% \input{\grammar/nrg.tex}

% Figure \ref{fig:grammar:nrg} shows an example of a derivation in an NRG to produce a node-labeled graph. Starting from the start symbol $S$, when a rule $\left(A\rightarrow R\right)$ is applied to a node $v$, the node together with edges linked to it is removed and the graph fragment $R$ is inserted. The embedding mechanism in $R$ includes node pairs associated with edge directions \citep{graphgrammar:Rozenberg:1997} to indicate how to add connections between neighbors of $v$ and nodes in R. 

% Similar to SERG, a \gls{snrg} which can be used in SMT is defined in Definition \ref{def:snrg}.

\begin{definition}
\label{def:snrg}
\normalfont A {\em synchronous node replacement grammar} (SNRG) is a tuple $\langle N,T,T',P,S\rangle$, where $N$ is a finite set of non-terminal symbols, $T$ and $T'$ are finite sets of terminal symbols, and $S\in N$ is the start symbol. $P$ is a finite set of productions of the form $\left(A\rightarrow\langle  R,R',\sim\rangle\right)$, where $A\in N$, $R$ is a graph fragment over $N\bigcup T$ and $R'$ is a graph fragment over $N\bigcup T'$. $\sim$ is a one-to-one mapping between non-terminal symbols in $R$ and $R'$.
\end{definition}

Note that the embedding mechanism is important during the generation of graphs, but can be ignored when parsing graphs \citep{graphgrammar:phdthesis:Kukluk:2014}. This is because generation requires adding connections between two graphs to form a new graph, whereas parsing graphs has no such requirements. Therefore, instead of using an SNRG exactly following Definition \ref{def:snrg}, we use a simplified version, which excludes embedding mechanisms, to build a translation model which parses source graphs and generates target strings. Informally, rules in our SNRG-based model are in the form of (\ref{eq:grammar:snrgrule}):
\begin{equation}\label{eq:grammar:snrgrule}
    X\to\langle\gamma,\alpha,\sim\rangle
\end{equation}
where $X$ is the general non-terminal, $\gamma$ is a graph where nodes are labeled by source terminals and non-terminals, $\alpha$ is a string over target terminals and non-terminals, and $\sim$ is a one-to-one mapping between source and target non-terminals. Figure \ref{fig:snrg:rule} shows an example translation rule.

\input{input/fig-snrgrule.tex}

\subsection{Rule Extraction}\label{sec:snrg:train}

In addition to rules which only contain terminals as in \seggbmt, \snrg~also includes translation rules which contain non-terminals as in Figure \ref{fig:snrg:rule}. Non-terminals are obtained by replacing subgraphs with single nodes and also can be replaced by subgraphs. Such a replacement requires representing a subgraph by joining other subgraphs. Because we only need to handle graphs on the source side, given a source subsequence $\tilde{s}$, the subgraph $G(\tilde{s})$ covering $\tilde{s}$ is already known and unique. Therefore, this subgraph-joining problem is simplified as a join of two subsequences, as defined in Definition \ref{def:join}.

\begin{definition}\label{def:join}
Given a sequence $s$, the {\em join} of its two disjoint subsequences $\tilde{s_1}$ of length $m$ and $\tilde{s_2}$ of length $n$ is a subsequence $\tilde{s}$ of length $m+n$ such that $\forall s_i: s_i\in\tilde{s}\Leftrightarrow s_i\in\tilde{s_1} \text{ or }\tilde{s_2}$. In this paper, the {\em join} is denoted as a commutative operation $\oplus$, i.e., $\tilde{s}=\tilde{s_1}\oplus\tilde{s_2}=\tilde{s_2}\oplus\tilde{s_1}$.
\end{definition}

According to Definition \ref{def:join}, word order is preserved during the joining of two subsequences. For example, the join of {\it shijibei juxing} and {\it zai Nanfei} is a subsequence {\it shijiebei zai Nanfei juxing}. It is trivial to keep word order for terminals. However, the existence of non-terminals brings another question: what is the position of a non-terminal $X$ covering a subsequence $\tilde{s_1}$ when it is joined with another subsequence $\tilde{s_2}$? It is straightforward to join $X$ with $\tilde{s_2}$ when spans of $\tilde{s_1}$ and $\tilde{s_2}$ do not overlap: if $s_1^b > s_2^e$, $X\oplus\tilde{s_2}=\tilde{s_2}X$; otherwise, if $s_1^e<s_2^b$, $X\oplus\tilde{s_2}=X\tilde{s_2}$. We now provide a definition of how to join $X$ with $\tilde{s_2}$ when the two spans overlap:

\begin{definition}\label{def:xpos}
The position of a non-terminal which covers a subsequence $\tilde{s}$ is the start position of $\tilde{s}$.
\end{definition}

For example, assuming $X$ covers {\it shijiebei juxing}, the join of $X$ with {\it zai nanfei} would result in {\em $X$ zai nanfei} as the start position of $X$, i.e. the position of {\it shijiebei}, is prior to the position of {\em zai}. This definition is useful when we extract rules and decode source graphs where we need to represent a subgraph with the non-terminal $X$.

Based on Definition \ref{def:join} and Definition \ref{def:xpos}, the set of rules is obtained in two steps by a similar extraction algorithm as in the HPB model, except that the source sides of rules in our model are graphs rather than strings. The rule set is defined over subgraph-phrase pairs (Definition \ref{def:spp}). Given a word-aligned graph--string pair $P=\langle G(s),t,a\rangle$, the set of rules from $P$ satisfies the following:
\begin{enumerate}
\item If $\langle G(\tilde{s}), \bar{t}\rangle$ is a subgraph--phrase pair, then
$$X\to\langle G(\tilde{s}),\bar{t}\rangle$$
is a rule of $P$.
\item If $X\to\langle\gamma,\alpha\rangle$ is a rule of $P$ and $\langle G(\tilde{s_1}),\bar{t}\rangle$ is a subgraph--phrase pair such that $\gamma=G(\tilde{s_1}\oplus\tilde{s_2})$ and $\alpha={r}_1 \overline{t}{r}_2$, then
$$X\to\langle G(\tilde{s_2}\oplus X_{[k]}),{r}_1 X_{[k]} {r}_2\rangle$$ is a rule of $P$, where $k$ is a unique index for a pair of non-terminal symbols.
\end{enumerate}

\input{input/fig-snrgextract.tex}

All \snrg~rules can be automatically learned from word-aligned graph--string pairs. A rule extractor firstly extracts rules without non-terminals which will be subsequently used to produce recursive rules by replacing subgraph-phrase pairs inside them with non-terminals. The extraction algorithm is similar to the one in the HPB model, except that we handle source subgraphs which have structures and may cover discontinuous phrases. Figure \ref{fig:snrg:train} illustrates how to extract a rule containing non-terminals. As in the HPB model, restrictions are added to the rule extractor to avoid generating a large volume of rules, namely:
\begin{enumerate}
    \item The length of subgraph-phrase pairs on both sides is limited to 10 at maximum.
    \item The number of symbols on the source side of a rule is limited to 5.
    \item Rules can have at most two non-terminals.
    \item There is at least one pair of aligned words in a rule.
\end{enumerate}

In addition to the translation rules above, two glue rules \citep{hpb:chiang:2005,hpb:chiang:2007} are used for robustness:
\begin{align}
    S&\to\langle S_{[1]}X_{[2]},S_{[1]}X_{[2]}\rangle \\
    S&\to\langle X_{[1]},X_{[1]}\rangle
\end{align}
Glue rules segment a graph into a sequence of subgraphs which will be translated separately, and then their translations are combined without reordering. They work similarly to glue rules in the HPB model. With the help of glue rules, we can make sure to obtain at least one translation of any input graph.

\subsection{Model and Features}\label{sec:snrg:model}

We define our model in the log-linear framework over a derivation $d=r_1r_2\cdots r_N$, as in Equation (\ref{eq:loglinear}). In our experiments, we use the standard 8 features in the HPB model: 
\begin{itemize}
    \item Two translation probabilities $p(G(s)|t)$ and $p(t|G(s))$;
    \item Two lexical translation probabilities $p_{\mathrm{lex}}(s|t)$ and $p_{\mathrm{lex}}(t|s)$;
    \item A language model $p(t)$ over a translation $t$;
    \item A rule penalty $\exp(-n)$ where $n$ is the number of non-glue rules;
    \item A word penalty $\exp(-|t|)$;
    \item A glue rule penalty $\exp(-m)$ where $m$ is the number of glue rules;
\end{itemize}
In addition, we add two new features:
\begin{itemize}
    \item A distortion penalty $\exp(-d(\cdot))$ as defined in Equation (\ref{eq:seggbmt:distortion}) when glue rules are used;
    \item A gap penalty $\exp(-g(d))$ where $g(d)$ is the total number of gaps introduced by non-glue rules in the derivation $d$.
\end{itemize}
Assuming a rule $r=\langle\gamma,\alpha\rangle$ translates a subsequence $\tilde{s}$ by replacing non-terminals with translations of smaller subsequences $\tilde{s_1},\cdots,\tilde{s_k}$, the number of gaps introduced by the rule can be calculated by Equation (\ref{eq:gap}).
\begin{equation}\label{eq:gap}
g(r) = (\tilde{s}^e-\tilde{s}^b+1) - |\{s_i\mid s_i\in\gamma\text{ or } \exists j:i\in [\tilde{s}_j^b,\tilde{s}_j^e]\}|
\end{equation}
Equation (\ref{eq:gap}) is a subtraction of two values. The first value is the span length of $\tilde{s}$ while the second value is the number of words which are considered by the current rule or by the gap penalty from previous rules. For example, given a rule which covers $\tilde{s}=s_1s_2s_3s_5s_8$ and has two non-terminals covering $\tilde{s_1}=s_1s_3$ and $\tilde{s_2}=s_2s_5$, respectively:
\begin{extract}
\begin{tabular}{ccccccccc}
    $s=$ & $s_1$ & $s_2$ & $s_3$ & $s_4$ & $s_5$ & $s_6$ & $s_7$ & $s_8$\\
    $\tilde{s}=$ & $s_1$ & $s_2$ & $s_3$ &  & $s_5$ &  &  & $s_8$\\
    $\tilde{s_1}=$ & $s_1$ &  & $s_3$ &  &  &  &  & \\
    $\tilde{s_2}=$ &  & $s_2$ &  &  & $s_5$ &  &  & \\
\end{tabular}
\end{extract}
According to Equation (\ref{eq:gap}), the span length of $\tilde{s}$ is $8-1+1=8$, while the number of words covered is $|\{s_1s_2s_3s_4s_5s_8\}|=6$. Therefore. the number of gaps introduced by the rules is $8-6=2$.

\subsection{Decoding}\label{sec:snrg:decode}

\input{input/fig-snrgderive.tex}

Similar to the \seggbmt, during decoding, the \snrg~model searches for the best derivation $\hat{d}$ whose source yield $f(\hat{d})$ corresponds to $G(s)$ and target yield $e(\hat{d})$ is a target sentence $\hat{t}$. Figure \ref{fig:snrg:derivation} shows a derivation which parses a Chinese DBG and simultaneously generates an English string. When a rule is applied, a subgraph in the source graph is replaced by a non-terminal node, and a new hypothesis is generated. Non-terminals in the target string of the rule are replaced by previous hypotheses.

In this section, following \citet{hpb:chiang:2007}, we present the decoding procedure for \snrg~as a deductive proof system \citep{proofsys:shieber:1995,proofsys:goodman:1999} which consists of (i) a set of weighted items $I:w$ containing axioms and goals, and (ii) a set of inference rules of the form:
\begin{displaymath}
\frac{I_1:w_1~\cdots~I_k:w_k}{I:w}~\Phi
\end{displaymath}
where items $I_i$ are {\em antecedents} with weights $w_i$, the item $I$ is a {\em consequence} with a weight $w$, and $\Phi$ is a side condition. The inference rule means that given all proven items $I_i:w_i$ and the side condition $\Phi$, we can derive $I:w$. Axioms are consequences without antecedents while goals are items which will cause the inference process to stop once proven. 

Similar to the HPB model, items in our deductive system can take one of two forms:
\begin{itemize}
    \item $[X,\tilde{s}]$ denoting that a subgraph or a sequence of subgraphs with a non-terminal $X$ and covering $\tilde{s}$ have already been recognized;
    \item $X\to\gamma$ if $(X\to\langle\gamma,\alpha\rangle)$ belongs to the SNRG \citep{proofsys:goodman:1999,hpb:chiang:2007}.
\end{itemize}
Note that we simply use the covered subsequence $\tilde{s}$ to represent a subgraph $G(\tilde{s})$ or a sequence of subgraphs $G(\tilde{s_1})\cdots G(\tilde{s_n})$ where $\tilde{s}=\tilde{s_1}\oplus\cdots\oplus\tilde{s_n}$. The sequence of subgraphs exists because glue rules combine subgraphs which may be disconnected. For simplicity, we use $\mathcal{G}(\tilde{s})$ to denote a sequence of subgraphs covering $\tilde{s}$. Clearly, $\mathcal{G}(\tilde{s})=G(\tilde{s})$ when $G(\tilde{s})$ exists. 

The inference process in the deductive system starts from axioms in the form of (\ref{eq:deductive:axiom}):
\begin{equation}\label{eq:deductive:axiom}
\frac{}
     {X\to\gamma:w} 
\quad (X\xrightarrow{w}\langle\gamma,\alpha\rangle)\in \text{ SNRG}
\end{equation}
Each rule in \snrg~with a weight $w$, including glue rules, is an axiom. The goal item in the system is $[S, s]$ which means that we have already recognized the source graph covering $s$. Given the axioms, our decoder can derive new weighted items in three ways:
\begin{itemize}
    \item If a rule consists of only terminals, we can use the following inference rule
    \begin{equation}\label{eq:deductive:term}
    \frac{X\to G(\tilde{s}):w}
         {[X,\tilde{s}]:w}
    \end{equation}
    to generate a new item $[X, \tilde{s}]$ with a weight $w$.
    
    \item Given a rule containing only one non-terminal $X$, if the deductive system has recognized a subsequence $\tilde{s_3}$ covered by the non-terminal, then we use the following rule to create a new weighted item:
    \begin{equation}\label{eq:deductive:1nt}
    \frac{Z\to G(\tilde{s_1}\oplus X\oplus\tilde{s_2}):w_1\quad [X,\tilde{s_3}]:w_2}
         {[Z,\tilde{s_1}\oplus\tilde{s_2}\oplus\tilde{s_3}]:w_1w_2}
    \end{equation}
    
    \item Finally, if we have a rule which contains two non-terminals, each of which has been recognized, we can use the following rule to derive a new weighted item.
    \begin{equation}\label{eq:deductive:2nt}
    \frac{Z\to \mathcal{G}(\tilde{s_1}\oplus X\oplus\tilde{s_2}\oplus Y\oplus\tilde{s_3}):w_1\quad [X,\tilde{s_4}]:w_2 \quad [Y,\tilde{s_5}]:w_3}
         {[Z,\tilde{s_1}\oplus\tilde{s_4}\oplus\tilde{s_2}\oplus\tilde{s_5}\oplus\tilde{s_3}]:w_1w_2w_3}
    \end{equation}
    Note that different from rule (\ref{eq:deductive:1nt}), rule (\ref{eq:deductive:2nt}) uses the notion of $\mathcal{G}(\tilde{s})$. This is because disconnectivity may appear only when glue rules are used to combine two subgraphs each of which has been recognized.
\end{itemize}

Inspired by the conventional chart decoder for SCFGs, ideally, proven items should be organized into a {\em chart} \citep{hpb:chiang:2007} where each cell $chart[X,\tilde{s}]$ consists of a set of items which have the same non-terminals and cover the same subsequences. However, because the number of subsequences in a sentences $s$ is exponential to the sentence length $|s|$, the chart would contain $2^{|s|}$ cells, which results in exponential time and space complexity so will not work in practice. Since the high complexity is caused by the exponential number of subsequences, to efficiently decode source graphs, we will add some restrictions so that the number of allowed subsequences can be reduced to be polynomial or even linear to sentence length.

\subsubsection{Beam-Search Decoder}\label{sec:snrg:decode:beam}

The first decoder we build is based on beam search inspired by the decoder in Section \ref{sec:seggbmt:decode}. This means that hypotheses which cover the same number of source words and have the same non-terminals are organized into the same stack. The difference is that the decoder in \snrg~generates translations in a bottom-up manner rather than left-to-right. In addition, hypotheses are grouped into different stacks according to their corresponding non-terminals. This is because glue rules (using non-terminals $S$) combine two subgraphs and their translations by ignoring the connectivity of the two subgraphs.

\input{input/alg-snrgdecoderbeam.tex}

A decoding procedure based on beam search for \snrg~is shown in Algorithm \ref{alg:snrgdecode:beam}. In the algorithm, proven items are organized into stacks $B[\cdot,\cdot]$, according to the number of covered source words and their non-terminals (Line 6 and Line 9), so that these items can be recombined and pruned according to their partial translation cost and an estimated future cost \citep{pb:koehn:2003,DTU:Galley:2010}. When the decoder has recognized all subsequences with length $l$, items which are in the same stack are grouped into {\it chart} for inferences in the next iteration (Lines 11--12). The translation ends when items which cover the whole sentence are proven. For efficiency and effectiveness, the maximum size of subgraphs is limited to a certain value $L_{\mathrm{max}}$ (20 in our experiments) (Line 4), and the maximum span of a subsequence covered by a subgraph is limited to 20 as well. Therefore, the total number of stacks in this algorithm is $L_{\mathrm{max}}+|s|$. The maximum number of subsequences allowed is reduced from exponential to $(L_{\mathrm{max}}+|s|)b$, where $b$ is the beam width.

\subsubsection{Chart Decoder}

Although the beam search decoder efficiently reduces the time and space complexity by organizing hypotheses which cover the same number of source words and have the same non-terminals into the same stack, it is known to suffer from search errors \citep{smt:Koehn:2010}. Therefore, in this section, we present a chart decoder.

It is easy to see that the large number of subsequences is caused by the free combination of words. Therefore, in our chart decoder, we try to use another restriction: only subgraphs which cover continuous spans are allowed. Items $[X,\tilde{s}]$ in the deductive system can then be represented in the form of $[X,i,j]$ where $i$ and $j$ are the start and end positions of $\tilde{s}$, respectively. Therefore, items which cover the same phrases will be organized into the same cell. Algorithm \ref{alg:snrgdecode:chart} shows a decoding procedure where the continuity restriction reduces the complexity of decoding from exponential time to cubic time as in tree-based models. Note that although Algorithm \ref{alg:snrgdecode:chart} looks the same as the decoding algorithm in HPB \citep{hpb:chiang:2007}, the source side $\gamma$ of a rule is a subgraph rather than a sequence. In addition, instead of accessing all continuous spans as in the HPB model, our decoder only accesses spans which are covered by subgraphs.

\input{input/alg-snrgdecoderchart.tex}

The continuity restriction also has a significant impact on the rule-extraction algorithm and feature functions. As described in Section \ref{sec:snrg:train}, while recursive rules are being extracted, we need to check whether a subsequence is nested in another subsequence. Time complexity of the checking is linear to the size of subsequences. However, if the continuity restriction is adopted, the checking would be in constant time as we only need to compare their start and end positions. In addition, values of the distortion penalty and gap penalty will always be 0 when the restriction is applied, resulting in 8 active features as in the HPB model.

\subsubsection{Language Model Integration and Pruning}

Given that the target sides of translation rules are hierarchical phrases, we use the same policy as that in the HPB model \citep{hpb:chiang:2007} to integrate a language model. $k$-best lists are generated using the algorithm in \citet{hpb:chiang:2007}. To speed up the decoding process, cube pruning \citep{hpb:chiang:2007} is used when the decoder derives new items from proved items.

%%%%%%%%%%%%%%%%%%%%%%%%%%%%%%%%%%%%%%%%%%

\section{Experiments}\label{sec:exp}

In this paper, we conduct large-scale experiments on two language pairs, ZH--EN and DE--EN.  The two language pairs have syntactically different word order, and thus recursive rules ought to be more helpful for phrase reordering. We experimented with 9 systems implemented in Moses \citep{moses:koehn:2007} using the same default configurations:
\begin{itemize}
    \item {\bf PBMT} and {\bf HPBMT} are phrase-based and HPB models, respectively.
    
    \item {\bf SDTU} extends the phrase-based model by allowing source discontinuous phrases \citep{DTU:Galley:2010}.

    \item {\bf TBMT} extends the phrase-based model by using treelets as the basic translation units. Different from \citep{treelet:quirk:2005}, translations in TBMT are generated from left to right using beam search.

    \item {\bf\seggbmt} is our graph-based model which translates graphs using graph segmentation. When inputs are trees, it is similar to TBMT except that during training TBMT takes all unaligned words on the source side into consideration while \seggbmt~only uses unaligned words on boundaries.
    
    \item {\bf Dep2Str} is an improved dependency tree-to-string model \citep{dep2str:xie:2011} which handles non-syntactic phrases by decomposing dependency structures \citep{dep2str:liangyou:2014}.
    
    \item {\bf\snrg\textsubscript{b}} is our SNRG-based model with a beam-search decoder.
    
    \item {\bf\snrg\textsubscript{c}} is our SNRG-based model with a chart decoder.
    
    \item {\bf SERG} is a SERG-based model which translates an edge-labeled dependency structure ({\it DEG} for short) into a target string \citep{serg:li:2015}. The system SERG is similar to \snrg\textsubscript{chart} when DSGs are used as inputs except that SERG uses edge-labeled graphs. By default, we only use the general non-terminal $X$ on both source and target sides.
    
\end{itemize}

While PBMT, TBMT, SDTU and \seggbmt~are segmentation-based models\footnote{It is well known that lexical reordering models (LRMs) can significantly improve phrase-based systems. However, according to \citet{DTU:Galley:2010}, the overall improvement brought by an LRM on both phrase-based and DTU systems is similar. The results in Table 4.7 in the PhD thesis of \citet{thesis:liangyou:2017} also show similar improvement by incorporating an LRM into \seggbmt. Therefore, we presume the observation also applies to this article and did not conduct experiments on LRMs.}, HPBMT, Dep2Str, SERG and \snrg\textsubscript{\it x}~use synchronous grammars. All of these systems are implemented in Moses with the same settings to enable fair comparisons. The number of hypotheses in a stack, i.e. beam width, is always limited to 200 in all systems.

\subsection{Data Sets and Experimental Setup}

\input{input/tab-data.tex}

Table \ref{tab:data} provides a summary of our corpora. The ZH--EN training corpus is from the LDC data, including LDC2002E18, LDC2003E07, LDC2003E14, LDC2004T07, the Hansards portion of LDC2004T08, and LDC2005T06. NIST 2002 (MT02) is taken as a development set to tune weights while NIST 2004 (MT04) and NIST 2005 (MT05) are used as test sets to evaluate systems. The Stanford Chinese word segmenter \citep{segmentor:chang:2008} is used to segment Chinese sentences into words. The Stanford dependency parser \citep{dependency:chang:2009} parses a Chinese sentence into a projective dependency tree. 

The DE--EN training corpus is from WMT 2014, including Europarl V7 \citep{europarl:Koehn:2005} and News Commentary. News-Test 2011 (WMT11) is taken as a development set while News-Test 2012 (WMT12) and News-Test 2013 (WMT13) are our test sets. We tokenize German sentences with scripts in Moses and use mate-tools to perform morphological analysis and parse the sentences \citep{matetool:bohnet:2010}. Then, MaltParser converts the parse results into projective dependency trees \citep{dependency:nivre:2005}.

Word alignment is performed by GIZA++ \citep{align:heuristic:och:2003} with the heuristic function \emph{grow-diag-final-and} \citep{wmsd:koehn:2005}. We use SRILM \citep{srilm:stolcke:2002} to train a 5-gram language model on the Xinhua portion of the English Gigaword corpus 5th edition with modified Kneser-Ney discounting \citep{knsmooth:chen:1996}. Batch MIRA \citep{mira:cherry:2012} is used to tune weights with a maximum iteration of 25. We report BLEU scores \citep{bleu:papineni:2002} and significance averaged over three MIRA runs \citep{multeval:Clark:2011}.

\subsection{Evaluation of Translation Quality}

\input{input/tab-mainscore.tex}

Table \ref{tab:mainscore} shows BLEU scores of all systems on all test sets. We found that while on average TBMT is comparable with PBMT, SDTU is significantly better than PBMT (+1.2 BLEU on ZH--EN and +0.4 on DE--EN on average). This is not surprising because SDTU takes both continuous and discontinuous phrases into consideration resulting in many more rules extracted and used. Even though rules in TBMT are more robust \citep{treelet:quirk:2005}, SDTU is better than it because TBMT discards a lot of continuous phrases which are not connected in dependency trees but could be important to phrase coverage and system performance.

\seggbmt~with dependency trees as inputs performs significantly worse than PBMT (-1.0 BLEU on ZH--EN and -1.5 on DE--EN on average) and TBMT (-1.1 BLEU on ZH--EN and -1.7 on DE--EN on average) as it uses significantly fewer rules. However, when sibling links are added to dependency trees resulting in DSGs, \seggbmt~is significantly improved (+2.0 BLEU on both ZH--EN and DE--EN on average compared to SegGBMT on trees). When DBGs are used, \seggbmt~achieves the best performance on average among all segmentation-based models. The improvement is mainly because extra links introduce many more rules into \seggbmt~and DBGs have more phrases connected than DSGs. Compared with SDTU which considers all possible discontinuous phrases, \seggbmt~with graphs is comparable to it on ZH--EN and significantly better on DE--EN. In addition, \seggbmt~uses significantly fewer rules than SDTU. In Table \ref{tab:mainscore}, we found that SDTU learns a huge model which contains more than twice as many rules compared to SegGBMT with DBG inputs.

Figure \ref{fig:gbmt:eg-pb} shows examples of translations where \seggbmt~successfully translated a Chinese collocation {\em Yu\ldots WuGuan} into {\em has nothing to do with}. By contrast, PBMT failed to catch the generalization since it only considers continuous phrases. Figure \ref{fig:gbmt:eg-treelet} shows examples of translations where TBMT translated a discontinuous phrase {\em Dui \ldots Zuofa} only to one word {\em on} and therefore an important target word {\em practice} was dropped. By contrast, bigram relations allowed our system \seggbmt~to translate a more proper phrase {\em De Zuofa} to {\em practice of}.

\input{input/eg-pb.tex}

\input{input/eg-treelet.tex}

Compared with PBMT, HPBMT is significantly better (+2.9 BLEU on ZH--EN and +1.1 on DE--EN on average) because of its capability of phrase reordering by using synchronous grammars. Both Dep2Str and SERG are significantly better than HPBMT on ZH--EN (+0.4 BLEU on average in terms of SERG) but worse on DE--EN (-0.3 BLEU on average in terms of SERG). We found that on DE--EN the number of rules in the two systems is dramatically reduced from 684M to around 95M, whereas the reduction on ZH--EN is less significant (from 388M to 84M/131M). This means that the two systems use more strict restriction on dependency trees when extracting rules on DE--EN. However, the fact that the two systems uses significantly fewer rules than HPBMT also suggests that linguistic structures are helpful in reducing model size.

Although \snrg\textsubscript{b} is better than \seggbmt (+0.4 BLEU on ZH--EN and +0.4 on DE--EN on average when DSG is used), it is significantly worse than HPBMT (-1.6 BLEU on ZH--EN and -0.2 on DE--EN on average when DSG is used). We presume that this is mainly because \snrg\textsubscript{b} uses beam search to reduce its search space where better hypotheses could be wrongly pruned. However, \snrg\textsubscript{b} is a meaningful trial towards a more general graph-based translation system as it allows hypotheses covering discontinuous source phrases when decoding using synchronous grammars. By contrast, \snrg\textsubscript{c} outperforms HPBMT when graphs are used (+0.3 BLEU on ZH--EN and +0.3 on DE--EN on average when DSG is used). \snrg\textsubscript{c} with tree inputs is comparable with HPBMT on both language pairs. This suggests that dependency trees effectively reduces model size without degrading translation quality when non-terminal rules are allowed. Compared with tree inputs, DSGs and DBGs bring consistently improvement as they introduce many more rules. We also found that in both \snrg\textsubscript{b} and \snrg\textsubscript{c}, while DBGs achieves the best performance on ZH--EN, DSGs are better on DE--EN. This may be caused by the fact that Chinese sentences have a larger mean dependency distance than German sentences \citep{depdist:eppler:2013} resulting sibling links less effective than bigram links on ZH--EN.

When structure information is added to rules, one concern is about the data sparsity issue, i.e., a single rule is refined into multiple rules with different structures. However, by comparing the number of rules extracted by HPBMT and \snrg\textsubscript{c} (with DBGs as inputs), we found the increase in the number of rules brought by graph structures is very small (specifically 6.96\% on ZH--EN and 8.19\% on DE--EN). In terms of \seggbmt~with DBGs as inputs, after excluding rules with discontinuous phrases (around 30\%), the number of rules approximates to that in PBMT. This means our methods do not cause severe issues of data sparsity.

% \input{input/eg-basic.tex}

% \input{input/eg-basicdetail.tex}

% Figure \ref{fig:grammar:eg-basic} shows translations of a Chinese sentence from the three systems. We found that both SERG and SNRG generated better translations than that of HPBMT. By tracking translation rules used during decoding, we found our systems used two rules to perform phrase reordering. Figure \ref{fig:grammar:eg-basicdetail} shows a phrase alignment produced by SERG and SNRG. The phrase alignment is obtained by applying two reordering rules which helped our systems perform better:
% \begin{displaymath}
% \begin{split}
%     X & \to\langle\text{Yin }X_{[1]}\text{ Bei }X_{[2]}, X_{[2]} \text{ for } X_{[1]}\rangle\\
%     X & \to\langle\text{Yu ShiJiuRi }X_{[1]}, X_{[1]}\text{ on 19 september}\rangle
% \end{split}
% \end{displaymath}
% Note that for simplicity the two rules are represented in the form of HPB rules and graph edges are ignored.

\subsection{Time Complexity of Decoding}

Decoding procedures implemented in segmentation-based models follow the beam search algorithm used in the phrase-based model. The decoder goes over each beam stack which stores hypotheses covering a specific number of source words. Each hypothesis in the beam stack is extended by applying a set of rules matching the input sentence which are usually called {\it translation options} and collected before decoding using efficient data structures and algorithms \citep{smt:Koehn:2010,DTU:Galley:2010}. Apparently, the time complexity of such a decoding procedure is $O(|s|\times|B|\times|R_s|)$, where $|s|$ is the length of an input sentence $s$, $|B|$ is the beam width and $R_s$ is the number of translation options at a given step. In PBMT, the number of translation options is linear to the sentence length as only continuous phrases with a bounded length are allowed. Therefore, the decoding complexity of PBMT can be rewrite as $O(|s|^2\times|B|)$. By considering a maximum distortion limit $d_{\text{max}}$ the complexity can be further reduced to $O(|s|\times|B|\times d_{\text{max}})$ because only a limited number of translation options is available at a given step. However, because the number of discontinuous phrases of an input sentence is exponential to the input length, the complexity of SDTU would be $O(|s|\times|B|\times|s|^{L_{\text{max}}})$ where $L_{\text{max}}$ is the maximum phrase length. The complexity can be further reduced to $O(|s|\times|B|\times C)$, where the constant $C \propto g_{\text{max}}^{L_{\text{max}}}\times d_{\text{max}}$, by using a maximum span $g_{\text{max}}$ and the distortion limit. Similar to SDTU, both TBMT and \seggbmt~have a time complexity of $O(|s|\times|B|\times C)$, however, with a smaller constant value than SDTU. This is because the connectivity constraint on tree and graph structures  greatly reduces the number of discontinuous phrases (See the number of rules in Table \ref{tab:mainscore}).

Synchronous grammar-based models, such as HPBMT, Dep2Str, SERG and \snrg\textsubscript{c}, in our experiments use the chart decoder as in Algorithm \ref{alg:snrgdecode:chart}. The algorithm maintains a beam stack for each continuous span of a source sentence, and translations of large spans are constructed by combining translations of smaller spans by applying rules with (at most 2 in experiments) non-terminals. The time complexity of the chart decoder is $O(|s|^2\times|R_s|\times|B|^2)$. However, because tree structures used by Dep2Str and SERG greatly reduce the number of available spans, the two systems run faster in practice than HPBMT. Compared to HPBMT, \snrg\textsubscript{c} needs additional time spent on matching graph edges, However, the time is a small constant as the number of edges in a subgraph is bounded in our experiments. When trees and DSGs are used as inputs, \snrg\textsubscript{c} runs faster. When DBGs are used as inputs, \snrg\textsubscript{c} takes more time to decode than HPB. \snrg\textsubscript{b} uses a beam search decoder described in Algorithm \ref{alg:snrgdecode:beam} which takes discontinuous source spans into consideration. The algorithm goes over each beam stack and apply rules over each subgraphs with a specific size and span (Section \ref{sec:snrg:decode:beam}). Therefore, the complexity of decoding is reduced from being exponential to $O(|s|^2\times g_{\text{max}}^{L_{\text{max}}}\times|R_s|\times|B|^2)$.

\subsection{Influence of Edge Labels}

\input{input/tab-edgelabel.tex}

Our graphs combine dependency relations with sequential relations, including bigram relations and sibling relations, to enable non-syntactic phrases. By default, the two kinds of relations are used without distinction. However, it would be interesting to see how edge types impact on translation performance. Therefore, we conducted further experiments where graph edges are labeled by link types: either dependency or sequential. Table \ref{tab:edgelabel} shows BLEU scores of \seggbmt~with DBG inputs and \snrg\textsubscript{c} with DSG inputs when edge types are taken into consideration (+EdgeLabel). The two systems are chosen because they achieve the best performance among the segmentation-based models and synchronous grammar-based models, respectively.

Results show that edge types do not improve our systems on the two language pairs overall. This is reasonable since we found adding edge labels to rules did not significantly increase the number of rules in our systems, as shown in Table \ref{tab:edgelabel}. This suggests that when a rule is matched with a subgraph, in most cases edge types are matched as well.

\subsection{Influence of Linguistic Non-terminals}

\input{input/tab-posnt.tex}

Similar to HPBMT, by default in our synchronous grammar-based models, we only use a general non-terminal symbol $X$ on both source and target sides. In this section, we conducted experiments to examine the impact of non-terminals on our models. The source non-terminal $X$ is replaced by non-terminals based on POS tags, which can be easily obtained as a by-product of dependency parsing.

The definition of a linguistic non-terminal for a subgraph follows \citet{head:junhui:2012} and \citet{serg:li:2015}. When a subgraph is connected by dependency relations, there must be one and only one node whose dependency head is not in the subgraph. We then denote the node as a {\em head} of the subgraph and simply use its POS tag as a non-terminal to represent the subgraph. For example, the head of Figure \ref{fig:dbgphrtype:b} is {\it zai} whose POS tag is {\it P}, so the non-terminal for Figure \ref{fig:dbgphrtype:b} is {\it P}. When a subgraph is disconnected and thus has two or more heads, we use a joint POS tag of these heads as a non-terminal. For example, because the heads of Figure \ref{fig:dbgphrtype:e} are {\it shijiebei} and {\it zai}, the non-terminal for the subgraph is {\it NT\_P}, where {\it NT} and {\it P} are POS tags of the two heads, respectively.

By default, during training each non-terminal covers at least two source words, i.e., the minimum size of gaps is 2 (denoted as MGS=2). This setting significantly reduces the number of non-terminals which are POS tags of single words and may influence translation performance. Therefore, we conducted two groups of experiments with MGS=2 and MGS=1, respectively.

Table \ref{tab:posnt} shows BLEU scores of systems when linguistic non-terminals are used. We found that when MGS=2, linguistic non-terminals have no significant impact on translation performance of systems except SERG on DE--EN. When MGS=1, we first found that both SERG and \snrg\textsubscript{c} are improved compared to their counterpart with MGS=2 (e.g., +0.3 BLEU on ZH--EN and +0.1 on DE--EN on average in terms of \snrg\textsubscript{c}). This may be caused by that more rules are extracted when MGS=1. In addition, when MGS=1, POS tags have more significant influence, especially on ZH--EN (+0.8 BLEU on average on both systems). Because MGS=1 means more rules with single POS tags as non-terminals are included, this suggests that these non-terminals are more useful on ZH--EN.

%%%%%%%%%%%%%%%%%%%%%%%%%%%%%%%%%%%%%%%%%%

\section{Conclusion}\label{sec:conclusion}

In this paper, we present novel graph-based translation models which translate source graphs into target strings. Graphs are built on top of dependency trees with extra links added to make non-syntactic phrases connected. The first model we introduce is based on graph segmentation which segments a graph into a sequence of subgraphs and generates translations by combining subgraph translations. Because the model is weak at phrase reordering, we further present a model based on a synchronous node replacement grammar to learn recursive translation rules. Experiments on Chinese--English and German--English show that our graph-based models significantly outperformed sequence- and tree-based baselines. We also found that edge labels have no significant impact on translation performance.

In future work, we would like to consider using other kinds of graphs, such as graphs representing feature structures which have proven to be a powerful tool for modeling morpho-syntactic aspects of natural languages \citep{featstruct:phdthesis:yvette:2011,featstruct:phdthesis:philip:2014} and investigate the impact of parsers' accuracy on translation quality when graphs are used. Recent progress on neural networks shows a promising way to perform MT with less feature engineering effort \citep{encoderdecoder:cho:2014,attention:bahdanau:2014}. However, how to use graphs in neural MT models is still an open problem. Therefore, we did not compare our models with neural models in this paper as we mainly focus on examining the effectiveness of graphs in MT. In future, it would  be interesting to explore how these kinds of graphs can be used in neural MT and how they impact on its translation performance. It would also be interesting to try our methods on more experimental settings, such as low resource translation which is more challenging for neural MT than SMT \citep{sixchallenge:koehn:2017} and would strength our methods.

%%%%%%%%%%%%%%%%%%%%%%%%%%%%%%%%%%%%%%%%%

\begin{acknowledgments}
This research has received funding from the European Union's Horizon 2020 research and innovation programme under grant agreement n\textordmasculine~645452 (QT21). The ADAPT Centre for Digital Content Technology is funded under the SFI Research Centres Programme (Grant 13/RC/2106) and is co-funded under the European Regional Development Fund. The authors thank all anonymous reviewers for their insightful comments and suggestions which greatly improve this paper.%insightful comments and suggestions.
\end{acknowledgments}

% added by the authors
% \clearpage

\starttwocolumn
\bibliography{ref}

\end{document}

%% file: input/fig-exampletree.tex
\begin{figure}
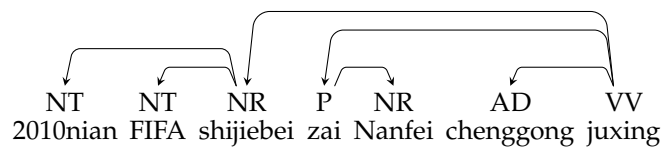

    \centering
    \begin{dependency}[hide label]
        \begin{deptext}[column sep=0cm]
        NT \& NT \& NR \& P \& NR \& AD \& VV\\
        2010nian \& FIFA \& shijiebei \& zai \& Nanfei \& chenggong \& juxing\\
        \end{deptext}
        
        % \deproot {7}{}
        \depedge {3}{1}{}
        \depedge {3}{2}{}
        \depedge {7}{3}{}
        \depedge {7}{4}{}
        \depedge {7}{6}{}
        \depedge {4}{5}{}
    \end{dependency}
    
    \caption{An example dependency tree of a Chinese sentence. Each node in the tree is labeled by a word associated with its Part-of-Speech (POS) tag.}
    \label{fig:tree}
\end{figure}

%% file: input/fig-examplegraph.tex
\begin{figure}
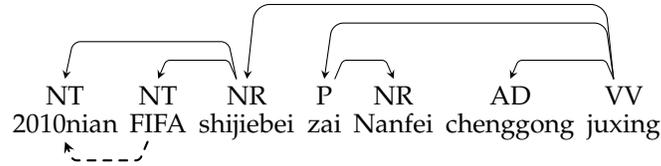

    \centering
    \begin{dependency}[hide label]
        \begin{deptext}[column sep=0cm]
        NT \& NT \& NR \& P \& NR \& AD \& VV\\
        2010nian \& FIFA \& shijiebei \& zai \& Nanfei \& chenggong \& juxing\\
        \end{deptext}
        
        \depedge {3}{1}{}
        \depedge {3}{2}{}
        \depedge {7}{3}{}
        \depedge {7}{4}{}
        \depedge {7}{6}{}
        \depedge {4}{5}{}
        \depedge [edge below, dashed, thick] {2}{1}{}        
    \end{dependency}
    
    \caption{An example graph of a Chinese sentence by adding an edge (dashed line) to the dependency tree in Figure \ref{fig:tree} so that the phrase {\it 2010nian FIFA} will be connected.}
    \label{fig:graph}
\end{figure}

%% file: input/fig-nodeorder.tex
\begin{figure}
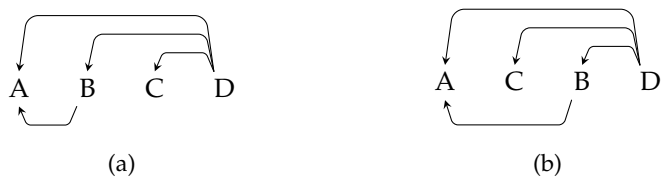

    \centering
    \begin{subfigure}[t]{0.4\linewidth}
        \centering
        \begin{dependency}[hide label]
            \begin{deptext}[column sep=5mm]
            A \& B \& C \& D\\
            \end{deptext}
            \depedge {4}{1}{}
            \depedge {4}{2}{}
            \depedge {4}{3}{}
            \depedge [edge below] {2}{1}{}        
        \end{dependency}
        \caption{}
        \label{fig:nodeorder:a}
    \end{subfigure}
    ~
    \begin{subfigure}[t]{0.4\linewidth}
        \centering
        \begin{dependency}[hide label]
            \begin{deptext}[column sep=5mm]
            A \& C \& B \& D\\
            \end{deptext}
            \depedge {4}{1}{}
            \depedge {4}{2}{}
            \depedge {4}{3}{}
            \depedge [edge below, edge unit distance=1ex] {3}{1}{}        
        \end{dependency}
        \caption{}
        \label{fig:nodeorder:b}
    \end{subfigure}
    
    \caption {Two example graphs. The two graphs are different because two nodes {\it B} and {\it C} are in different order: while (a) covers a sequence {\it A\underline{BC}D}, (b) covers {\it A\underline{CB}D}.}
    \label{fig:nodeorder}
\end{figure}

%% file: input/fig-subgraph.tex
\begin{figure}
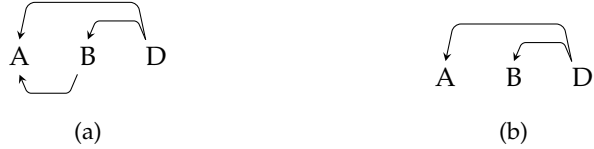

    \centering
    
    \begin{subfigure}[t]{0.4\linewidth}
        \centering
        \begin{dependency}[hide label]
            \begin{deptext}[column sep=5mm]
            A \& B \& D\\
            \end{deptext}
            \depedge {3}{1}{}
            \depedge {3}{2}{}
            \depedge [edge below] {2}{1}{}        
        \end{dependency}
        \caption{}
        \label{fig:subgraph:a}
    \end{subfigure}
    ~
    \begin{subfigure}[t]{0.4\linewidth}
        \centering
        \begin{dependency}[hide label]
            \begin{deptext}[column sep=5mm]
            A \& B \& D\\
            \end{deptext}
            \depedge {3}{1}{}
            \depedge {3}{2}{}      
        \end{dependency}
        \caption{}
        \label{fig:subgraph:b}
    \end{subfigure}
    
    \caption {Illustration of node-induced subgraphs. (a) is a node-induced subgraph of Figure \ref{fig:nodeorder:a} while (b) is not because it lacks an edge from B to A.}
    \label{fig:subgraph}
\end{figure}

%% file: input/fig-dbg.tex
\begin{figure}
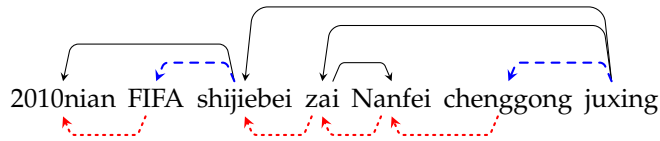

    \centering
    
    \begin{dependency}[hide label]
        \begin{deptext}[column sep=0cm]
        2010nian \& FIFA \& shijiebei \& zai \& Nanfei \& chenggong \& juxing\\
        \end{deptext}

        \depedge {3}{1}{}
        \depedge [dashed,blue,thick] {3}{2}{}
        \depedge {7}{3}{}
        \depedge {7}{4}{}
        \depedge [dashed,blue,thick] {7}{6}{}
        \depedge {4}{5}{}
        
        \depedge [edge below,color=red,dotted,thick] {2}{1}{}
        \depedge [edge below, color=red,dotted,thick] {4}{3}{}
        \depedge [edge below, color=red,dotted,thick] {5}{4}{}
        \depedge [edge below, color=red,dotted,thick] {6}{5}{}
    \end{dependency}
    \caption{An example dependency-bigram graph. Dotted lines are bigram relations. Solid lines are dependency relations. Dashed lines are shared by bigram and dependency relations.}
    \label{fig:dbg}
\end{figure}

%% file: input/fig-dbgphrtype.tex
\begin{figure}
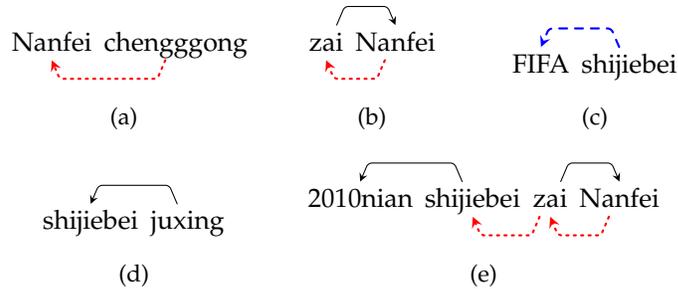

    \centering
    
    \begin{subfigure}[b]{0.25\linewidth}
    \centering
    \begin{dependency} [hide label]
        \begin{deptext}[column sep=0cm]
        Nanfei \& chengggong\\
        \end{deptext}
        \depedge [edge below, dotted, red, thick]{2}{1}{}
    \end{dependency}
    \caption{}
    \label{fig:dbgphrtype:a}
    \end{subfigure}
    ~
    \begin{subfigure}[b]{0.2\linewidth}
    \centering
    \begin{dependency} [hide label]
        \begin{deptext}[column sep=0cm]
        zai \& Nanfei\\
        \end{deptext}
        \depedge {1}{2}{}
        \depedge [edge below, dotted, red, thick]{2}{1}{}
    \end{dependency}
    \caption{}
    \label{fig:dbgphrtype:b}
    \end{subfigure}
    ~
    \begin{subfigure}[b]{0.2\linewidth}
    \centering
    \begin{dependency} [hide label]
        \begin{deptext}[column sep=0cm]
        FIFA \& shijiebei\\
        \end{deptext}
        \depedge [dashed,blue,thick] {2}{1}{}
    \end{dependency}
    \caption{}
    \label{fig:dbgphrtype:c}
    \end{subfigure}
    \\[0.1cm]    
    \begin{subfigure}[b]{0.25\linewidth}
    \centering
    \begin{dependency} [hide label]
        \begin{deptext}[column sep=0cm]
        shijiebei \& juxing\\
        \end{deptext}
        \depedge{2}{1}{}
    \end{dependency}
    \caption{}
    \label{fig:dbgphrtype:d}
    \end{subfigure}
    ~
    \begin{subfigure}[b]{0.4\linewidth}
    \centering
    \begin{dependency} [hide label]
        \begin{deptext}[column sep=0cm]
        2010nian \& shijiebei \& zai \& Nanfei\\
        \end{deptext}
        \depedge{2}{1}{}
        \depedge[edge below, red, dotted, thick]{3}{2}{}
        \depedge[edge below, red, dotted, thick]{4}{3}{}
        \depedge{3}{4}{}
    \end{dependency}
    \caption{}
    \label{fig:dbgphrtype:e}
    \end{subfigure}
    
    \caption{Example subgraphs of Figure \ref{fig:dbg}. Dotted lines are bigram relations. Solid lines are dependency relations. Dashed lines are shared by bigram and dependency relations. (a) is only connected by bigram links; (b) is connected by not only bigram links but also dependency links; (c) is connected by a link shared by bigram relations and dependency relations; (d) is only connected by dependency links; and (e) is connected when both bigram links and dependency links are used.}
    \label{fig:dbgphrtype}
\end{figure}

%% file: input/fig-dsg.tex
\begin{figure}
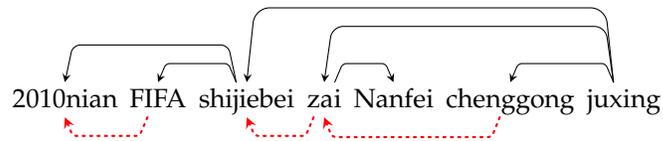

    \centering
    \begin{dependency}[hide label]
        \begin{deptext}[column sep=0cm]
        2010nian \& FIFA \& shijiebei \& zai \& Nanfei \& chenggong \& juxing\\
        \end{deptext}

        \depedge {3}{1}{}
        \depedge {3}{2}{}
        \depedge {7}{3}{}
        \depedge {7}{4}{}
        \depedge {7}{6}{}
        \depedge {4}{5}{}
        
        \depedge [edge below,color=red,dotted,thick] {2}{1}{}
        \depedge [edge below, color=red,dotted,thick] {4}{3}{}
        \depedge [edge below, color=red,dotted,thick,edge unit distance=0.8ex] {6}{4}{}
    \end{dependency}
    
    \caption{An example dependency-sibling graph by directly adding edges between siblings to a dependency tree. Dotted lines are sibling relations. Solid lines are dependency relations.}
    \label{fig:dsg}
\end{figure}

%% file: input/fig-dsgphrtype.tex
\begin{figure}
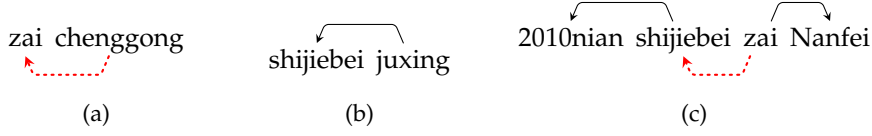

    \centering
    
    \begin{subfigure}[b]{0.24\linewidth}
    \centering
    \begin{dependency} [hide label]
        \begin{deptext}[column sep=0cm]
        zai \& chenggong\\
        \end{deptext}
        \depedge [edge below, dotted, red, thick]{2}{1}{}
    \end{dependency}
    \caption{}
    \label{fig:dsgphrtype:a}
    \end{subfigure}
    ~
    \begin{subfigure}[b]{0.24\linewidth}
    \centering
    \begin{dependency} [hide label]
        \begin{deptext}[column sep=0cm]
        shijiebei \& juxing\\
        \end{deptext}
        \depedge{2}{1}{}
    \end{dependency}
    \caption{}
    \label{fig:dsgphrtype:b}
    \end{subfigure}
    ~
    \begin{subfigure}[b]{0.38\linewidth}
    \centering
    \begin{dependency} [hide label]
        \begin{deptext}[column sep=0cm]
        2010nian \& shijiebei \& zai \& Nanfei\\
        \end{deptext}
        \depedge{2}{1}{}
        \depedge[edge below, red, dotted, thick]{3}{2}{}
        \depedge{3}{4}{}
    \end{dependency}
    \caption{}
    \label{fig:dsgphrtype:c}
    \end{subfigure}
    
    \caption{Example subgraphs of Figure \ref{fig:dsg}. Dotted lines are sibling relations. Solid lines are dependency relations. (a) is connected by sibling links; (b) is connected by dependency links; and (c) is connected when both types of links are considered.}
    \label{fig:dsgphrtype}
\end{figure}

%% file: input/fig-dbg2str.tex
\begin{figure}
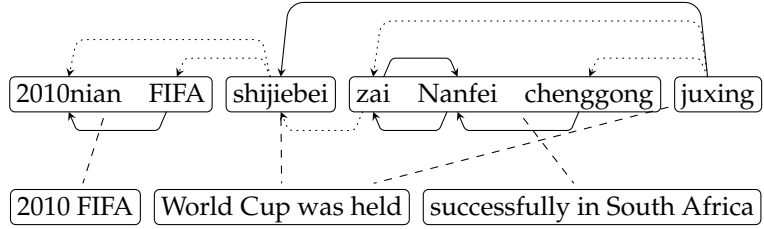

    \centering
    \begin{dependency}[hide label]
        \begin{deptext} [column sep=0.2cm]
        2010nian \& FIFA \& shijiebei \& zai \& Nanfei \& chenggong \& juxing\\
        \end{deptext}

        \depedge [dotted] {3}{1}{};
        \depedge [dotted] {3}{2}{};
        \depedge {7}{3}{};
        \depedge [dotted] {7}{4}{};
        \depedge [dotted] {7}{6}{};
        \depedge {4}{5}{};
        
        \depedge [edge below] {2}{1}{};
        \depedge [dotted, edge below] {4}{3}{};
        \depedge [edge below] {5}{4}{};
        \depedge [edge below] {6}{5}{};

        \wordgroup{1}{1}{2}{s1};
        \wordgroup{1}{3}{3}{s2};
        \wordgroup{1}{4}{6}{s3};
        \wordgroup{1}{7}{7}{s4};

        \begin{deptext} [yshift=-1.5cm, column sep=0.2cm]
        2010 FIFA \& World Cup was held \& successfully in South Africa \\
        \end{deptext}
        \wordgroup{1}{1}{1}{t1};
        \wordgroup{1}{2}{2}{t2};
        \wordgroup{1}{3}{3}{t3};

        \draw [dashed] (s1) -- (t1);
        \draw [dashed] (s2) -- (t2);
        \draw [dashed] (s3) -- (t3);
        \draw [dashed] (s4) -- (t2);
        
    \end{dependency}

    \caption {A source DBG is segmented into three subgraphs, each of which corresponds to a target phrase. Dashed lines denote alignments between source subgraphs and target phrases. Edges in dotted lines are ignored during segmention of the graph.}
    \label{fig:dbg2str}
\end{figure}

%% file: input/fig-subgraphphrase.tex
\begin{figure}
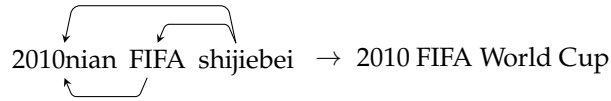

    \centering
    \begin{dependency}[hide label]
        \begin{deptext}
        2010nian \& FIFA \& shijiebei\\
        \end{deptext}
        \depedge[edge below]{2}{1}{}
        \depedge{3}{1}{}
        \depedge{3}{2}{}
        
        \node(t)[right = 0cm and 0.5cm of \matrixref] {2010 FIFA World Cup};
        \path(\matrixref) --node{$\to$} (t);
    \end{dependency}
    \caption{An example subgraph--phrase pair extracted from the example in Section \ref{sec:introduction} using the DBG in Figure \ref{fig:dbg}.}
    \label{fig:subgraphphrase}
\end{figure}

%% file: input/fig-sppair.tex
\begin{figure}
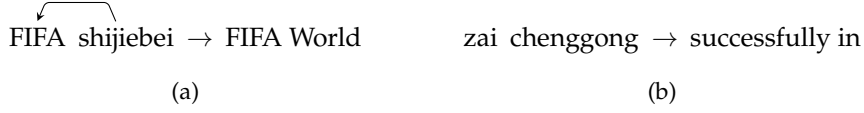

    \centering
    
    \begin{subfigure}[b]{0.45\linewidth}
    \centering
    \begin{dependency} [hide label]
        \begin{deptext}[column sep=0cm]
        FIFA \& shijiebei \& $\to$ \& FIFA World\\
        \end{deptext}
        \depedge{2}{1}{}
    \end{dependency}
    \caption{}
    \label{fig:invalidsppair:a}
    \end{subfigure}
    ~
    \begin{subfigure}[b]{0.45\linewidth}
    \centering
    \begin{dependency} [hide label]
        \begin{deptext}[column sep=0cm]
        zai \& chenggong \& $\to$ \& successfully in\\
        \end{deptext}
    \end{dependency}
    \caption{}
    \label{fig:invalidsppair:b}
    \end{subfigure}
    
    \caption{Examples pairs which are not subgraph-phrase pairs because: (a) is not consistent with the word alignment as {\it shijiebei} should be aligned to {\it World Cup}, and the source side of (b) is not a subgraph of the DBG in Figure \ref{fig:dbg}.}
    \label{fig:invalidsppair}
\end{figure}

%% file: input/fig-seggbmt-distortion.tex
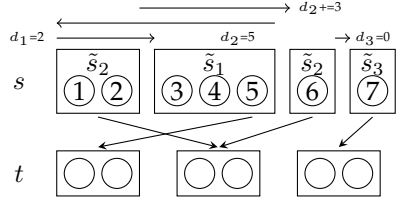
\begin{figure}
    \centering
    \begin{tikzpicture}
    
    \node at (-0.5, 0.4) {$s$};
    \draw (0,0) rectangle (1.1,0.85); \node at (0.55, 0.65) {$\tilde{s}_2$};
    \draw (1.3,0) rectangle (2.9,0.85); \node at (2.1, 0.65) {$\tilde{s}_1$};
    \draw (3.1,0) rectangle (3.7,0.85); \node at (3.4, 0.65) {$\tilde{s}_2$};
    \draw (3.9,0) rectangle (4.5,0.85); \node at (4.2, 0.65) {$\tilde{s}_3$};
    \draw (0.3, 0.3) circle [radius=0.2]; \node at (0.3,0.3) {1};
    \draw (0.8, 0.3) circle [radius=0.2]; \node at (0.8,0.3) {2};
    \draw (1.6, 0.3) circle [radius=0.2]; \node at (1.6,0.3) {3};
    \draw (2.1, 0.3) circle [radius=0.2]; \node at (2.1,0.3) {4};
    \draw (2.6, 0.3) circle [radius=0.2]; \node at (2.6,0.3) {5};
    \draw (3.4, 0.3) circle [radius=0.2]; \node at (3.4,0.3) {6};
    \draw (4.2, 0.3) circle [radius=0.2]; \node at (4.2,0.3) {7};
    
    \node at (-0.5, -0.8) {$t$};
    \draw (0,-1.1) rectangle (1.1, -0.5);
    \draw (1.6,-1.1) rectangle (2.7,-0.5);
    \draw (3.2,-1.1) rectangle (4.3,-0.5);
    \draw (0.3, -0.8) circle [radius=0.2];
    \draw (0.8, -0.8) circle [radius=0.2];
    \draw (1.9, -0.8) circle [radius=0.2];
    \draw (2.4, -0.8) circle [radius=0.2];
    \draw (3.5, -0.8) circle [radius=0.2];
    \draw (4, -0.8) circle [radius=0.2];
    
    \draw [-{stealth}] (2.6,-0.05) -- (0.55,-0.45);
    \draw [-{stealth}] (0.55,-0.05) -- (2.14,-0.45);
    \draw [-{stealth}] (3.4,-0.05) -- (2.16,-0.45);
    \draw [-{stealth}] (4.2,-0.05) -- (3.75,-0.45);
    
    \draw [<-] (1.3,1.0) -- (0,1.0); \node at (-0.4,1.0) {\tiny $d_1$=2};
    \draw [->] (2.9,1.2) -- (0.0,1.2); \node at (2.4,1.0) {\tiny $d_2$=5};
    \draw [->] (1.1,1.4) -- (3.1,1.4); \node at (3.5,1.4) {\tiny $d_2$+=3};
    \draw [->] (3.7,1.0) -- (3.9,1.0); \node at (4.2,1.0) {\tiny $d_3$=0};
    
    \end{tikzpicture}
    \caption {Distortion calculation in SegGBMT for both continuous ($\tilde{s}_1$ and $\tilde{s}_3$) and discontinuous ($\tilde{s}_2$) phrases in a derivation.}
    \label{fig:seggbmt:distortion}
\end{figure}

%% file: input/fig-seggbmt-decode.tex
\begin{figure}[t]
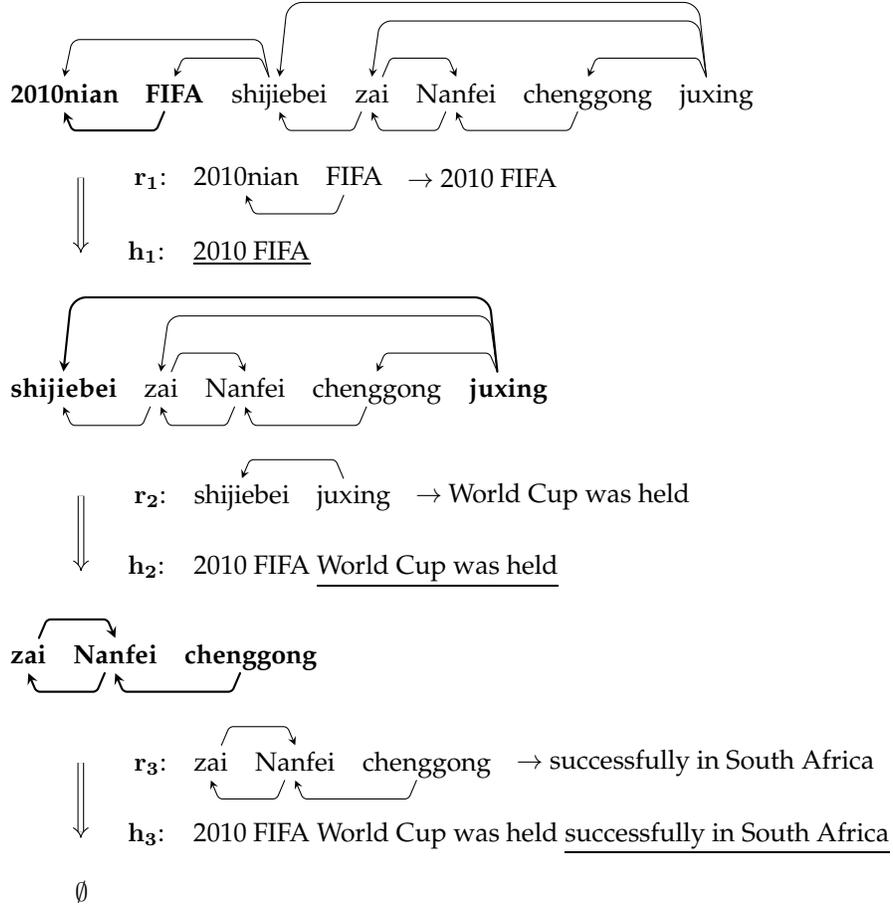

    \centering
    \begin{tabular}{l}
        
        \begin{dependency}[hide label,edge unit distance=1.5ex]
            \begin{deptext}[column sep=0.2cm]
            \bf 2010nian \& \bf FIFA \& shijiebei \& zai \& Nanfei \& chenggong \& juxing\\
            \end{deptext}
            \depedge {3}{1}{}
            \depedge {3}{2}{}
            \depedge {7}{3}{}
            \depedge {7}{4}{}
            \depedge {7}{6}{}
            \depedge {4}{5}{}
            \depedge [edge below, thick] {2}{1}{}
            \depedge [edge below] {4}{3}{}
            \depedge [edge below] {5}{4}{}
            \depedge [edge below] {6}{5}{}
            %
            % \node [left = 0cm and 0cm of \matrixref] {\it Input:};
        \end{dependency}
        \\
        \hspace{1cm}
        \begin{dependency}[hide label,edge unit distance=1.5ex]
            \begin{deptext}[column sep=0.2cm]
            $\mathbf{r_1}$: \& 2010nian \& FIFA \\
            \end{deptext}
            \depedge [edge below] {3}{2}{}
            
            \node (t) [right =0 and 0 of \matrixref] {$\to$ 2010 FIFA};
            % \draw [arrow,dashed,thick] (\matrixref) -- (t);
            \node (h) [below =of \matrixref.west,anchor=west] (h) {$\mathbf{h_1}$:\hspace{0.3cm} \underline{2010 FIFA}};
            
            \draw [-implies,double equal sign distance] ([xshift=-0.5cm]\matrixref.west) -- ([xshift=-0.5cm]h.west);
        \end{dependency}
        \\
        \begin{dependency}[hide label,edge unit distance=1.5ex]
            \begin{deptext}[column sep=0.2cm]
            \bf shijiebei \& zai \& Nanfei \& chenggong \& \bf juxing\\
            \end{deptext}
            \depedge [thick] {5}{1}{}
            \depedge {5}{2}{}
            \depedge {5}{4}{}
            \depedge {2}{3}{}
            \depedge [edge below] {2}{1}{}
            \depedge [edge below] {3}{2}{}
            \depedge [edge below] {4}{3}{}
            %
            % \node [left = 0cm and 0cm of \matrixref] {\it Input:};
        \end{dependency}
        \\
        \hspace{1cm}
        \begin{dependency}[hide label,edge unit distance=1.5ex]
            \begin{deptext}[column sep=0.2cm]
            $\mathbf{r_2}$: \& shijiebei \& juxing\\
            \end{deptext}
            \depedge {3}{2}{}
            
            \node (t) [right =0 and 0 of \matrixref] {$\to$ World Cup was held};
            % \draw [arrow,dashed,thick] (\matrixref) -- (t);
            \node (h) [below =of \matrixref.west,anchor=west] (h) {$\mathbf{h_2}$:\hspace{0.3cm} 2010 FIFA \underline{World Cup was held}};
            
            \draw [-implies,double equal sign distance] ([xshift=-0.5cm]\matrixref.west) -- ([xshift=-0.5cm]h.west);
        \end{dependency}
        \\
        \begin{dependency}[hide label,edge unit distance=1.5ex]
            \begin{deptext}[column sep=0.2cm]
            \bf zai \& \bf Nanfei \& \bf chenggong\\
            \end{deptext}
            \depedge [thick] {1}{2}{}
            \depedge [edge below,thick] {2}{1}{}
            \depedge [edge below, thick] {3}{2}{}
            
            % \node [left = 0cm and 0cm of \matrixref] {\it Input:};
        \end{dependency}
        \\
        \hspace{1cm}
        \begin{dependency}[hide label,edge unit distance=1.5ex]
            \begin{deptext}[column sep=0.2cm]
            $\mathbf{r_3}$: \& zai \& Nanfei \& chenggong\\
            \end{deptext}
            \depedge {2}{3}{}
            \depedge [edge below] {3}{2}{}
            \depedge [edge below] {4}{3}{}
            
            \node (t) [right =0 and 0 of \matrixref] {$\to$ successfully in South Africa};
            % \draw [arrow,dashed,thick] (\matrixref) -- (t);
            \node (h) [below =of \matrixref.west,anchor=west] (h) {$\mathbf{h_3}$:\hspace{0.3cm} 2010 FIFA World Cup was held \underline{successfully in South Africa}};
            
            \draw [-implies,double equal sign distance] ([xshift=-0.5cm]\matrixref.west) -- ([xshift=-0.5cm]h.west);
        \end{dependency}
        \\
        \hspace{0.85cm}
        \begin{dependency}
            \node(a){$\emptyset$};
        \end{dependency}
    \end{tabular}
    
    \caption [A derivation of translating an dependency-bigram graph] {A derivation of translating a DBG. Each rule $\mathbf{r_i}$ matches an input subgraph (in bold) and generates a new hypothesis $\mathbf{h_i}$ by appending translations (underlined) of the subgraph to the right.}
    \label{fig:seggbmt:decode}
\end{figure}

%% file: input/alg-seggbmtdecoder.tex
\begin{algorithm}
\caption{Beam-search decoder for SegGBMT}
\label{alg:seggbmt:decode}

\begin{algorithmic}[1]
    \State add $h_\emptyset$ to $B_0$
    \For{$i=0$ to $|s|$}
        \ForAll{$h_{c} \in B_i$}
            \State $j=\min\{k\mid k\notin c\}$\Comment{the first uncovered position}
            \ForAll{$r\in\{\langle G(\tilde{s}), \bar{t}\rangle\}: \tilde{s}^b\le j+d_{\text{max}} \text{ and } \forall k\in c\Rightarrow s_k\notin\tilde{s}$}
                % \If{}
                    \State $h_{c'}:=\mathrm{Create}(h_c, r)$ where $c'=c\cup\{k\mid s_k\in\tilde{s}\}$
                    \State add $h_{c'}$ to $B_{i+|\tilde{s}|}$
                    \State recombine and prune if applicable
                % \EndIf
            \EndFor
        \EndFor
    \EndFor
    % \State \Return $\argmax_{h_c}B_{|s|}$
\end{algorithmic}

\end{algorithm}

%% file: input/fig-graphfrag.tex
\begin{figure}
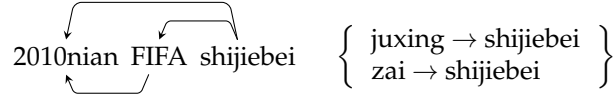

    \centering
    \begin{dependency}[hide label]
        \begin{deptext}
        2010nian \& FIFA \& shijiebei\\
        \end{deptext}
        \depedge[edge below]{2}{1}{}
        \depedge{3}{1}{}
        \depedge{3}{2}{}
        
        \node(t)[right = 0cm and 0.2cm of \matrixref] {
            \( \left \{
            \begin{tabular}{l}
                 juxing $\to$ shijiebei \\
                 zai $\to$ shijiebei \\
            \end{tabular}
            \right \} \)
        };
    \end{dependency}
    \caption{An example graph fragment which consists of two parts: a graph (on the left) and an embedding mechanism (on the right) which contains two links. When we integrate the graph into another one, if {\it juxing} and {\it zai} are neighbours of the graph, the two links will be added.}
    \label{fig:graphfrag}
\end{figure}

%% file: input/fig-snrgrule.tex
\begin{figure}
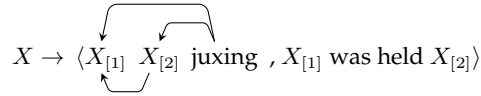

    \centering
    \small
    
    \begin{dependency}[hide label]
        \begin{deptext}
       $X\to$ \& $\langle X_{[1]}$ \& $X_{[2]}$ \& juxing \& , $X_{[1]}$ was held $X_{[2]}\rangle$\\
        \end{deptext}
        \depedge[edge below]{3}{2}{}
        \depedge{4}{2}{}
        \depedge{4}{3}{}
    \end{dependency}
    
    \caption {An example translation rule in our SNRG-based model. $X$ is a general non-terminal. Indexes indicate mappings between source and target non-terminals. }
    \label{fig:snrg:rule}
\end{figure}

%% file: input/fig-snrgextract.tex
\begin{figure}
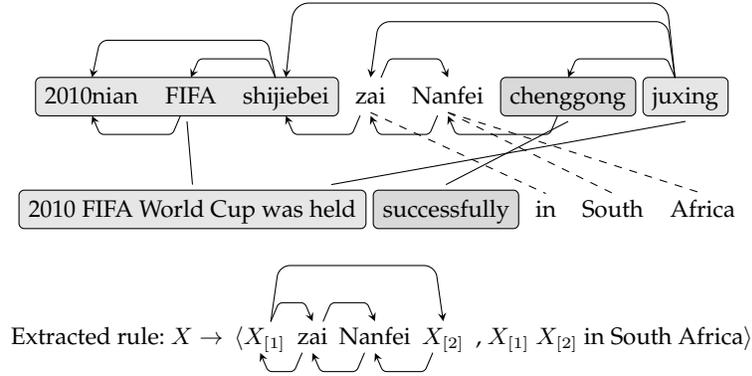

    \centering
    \small
    \begin{dependency}[hide label, edge vertical padding=.3ex]
    
    %%%%% source side
        \begin{deptext}[column sep=0.2cm]
        2010nian \& FIFA \& shijiebei \& zai \& Nanfei \& chenggong \& juxing\\
        \end{deptext}
        \depedge {3}{1}{}
        \depedge {3}{2}{}
        \depedge {7}{3}{}
        \depedge {7}{4}{}
        \depedge {7}{6}{}
        \depedge {4}{5}{}
        \depedge [edge below] {2}{1}{}
        \depedge [edge below] {4}{3}{}
        \depedge [edge below] {5}{4}{}
        \depedge [edge below] {6}{5}{}
        \wordgroup[group style={fill=black!10, inner sep=.3ex}]{1}{1}{3}{sp1}
        \wordgroup[group style={fill=black!10, inner sep=.3ex}]{1}{7}{7}{sp2}
        \wordgroup[group style={fill=black!15, inner sep=.3ex}]{1}{6}{6}{sp3}
        
        \coordinate (s1) at (\wordref{1}{1}.south);
        \coordinate (s2) at (\wordref{1}{2}.south);
        \coordinate (s3) at (\wordref{1}{3}.south);
        \coordinate (s4) at (\wordref{1}{4}.south);
        \coordinate (s5) at (\wordref{1}{5}.south);
        \coordinate (s6) at (\wordref{1}{6}.south);
        \coordinate (s7) at (\wordref{1}{7}.south);
    
    %%%%%%% target side
        
        \begin{deptext}[column sep=0.2cm, yshift=-1.5cm]
        2010 FIFA World Cup was held \& successfully \& in \& South \& Africa\\
        \end{deptext}
        \wordgroup[group style={fill=black!10, inner sep=.3ex}]{1}{1}{1}{tp1}
        \wordgroup[group style={fill=black!15, inner sep=.3ex}]{1}{2}{2}{tp2}
        \coordinate (t1) at (\wordref{1}{1}.north);
        \coordinate (t2) at (\wordref{1}{2}.north);
        \coordinate (t3) at (\wordref{1}{3}.north);
        \coordinate (t4) at (\wordref{1}{4}.north);
        \coordinate (t5) at (\wordref{1}{5}.north);
    
    %%%%% alignment
        \draw (sp1.south) -- (tp1.north);
        \draw (sp2.south) -- (tp1);
        \draw (sp3.south) -- (tp2.north);
        \draw [dashed] (s4) -- (t3.north);
        \draw [dashed] (s5) -- (t4.north);
        \draw [dashed] (s5) -- (t5.north);
        
    %%%%% extracted rule
        \begin{deptext}[yshift=-3.2cm]
        Extracted rule: $X\to$ \& $\langle X_{[1]}$ \& zai \& Nanfei \& $X_{[2]}$ \& , $X_{[1]}$ $X_{[2]}$ in South Africa$\rangle$\\
        \end{deptext}
        \depedge[edge vertical padding=0]{2}{3}{}
        \depedge[edge vertical padding=0]{2}{5}{}
        \depedge[edge vertical padding=0]{3}{4}{}
        \depedge[edge below,edge vertical padding=0]{3}{2}{}
        \depedge[edge below,edge vertical padding=0]{4}{3}{}
        \depedge[edge below,edge vertical padding=0]{5}{4}{}
        
    \end{dependency}
    
    \caption{Illustrating the extraction of a translation rule in \snrg~by replacing subsequences in grey with non-terminals. Indexes on non-terminals indicate mappings. Solid lines are phrase alignment while dashed lines are word alignment.}
    \label{fig:snrg:train}
\end{figure}

%% file: input/fig-snrgderive.tex
\begin{figure}
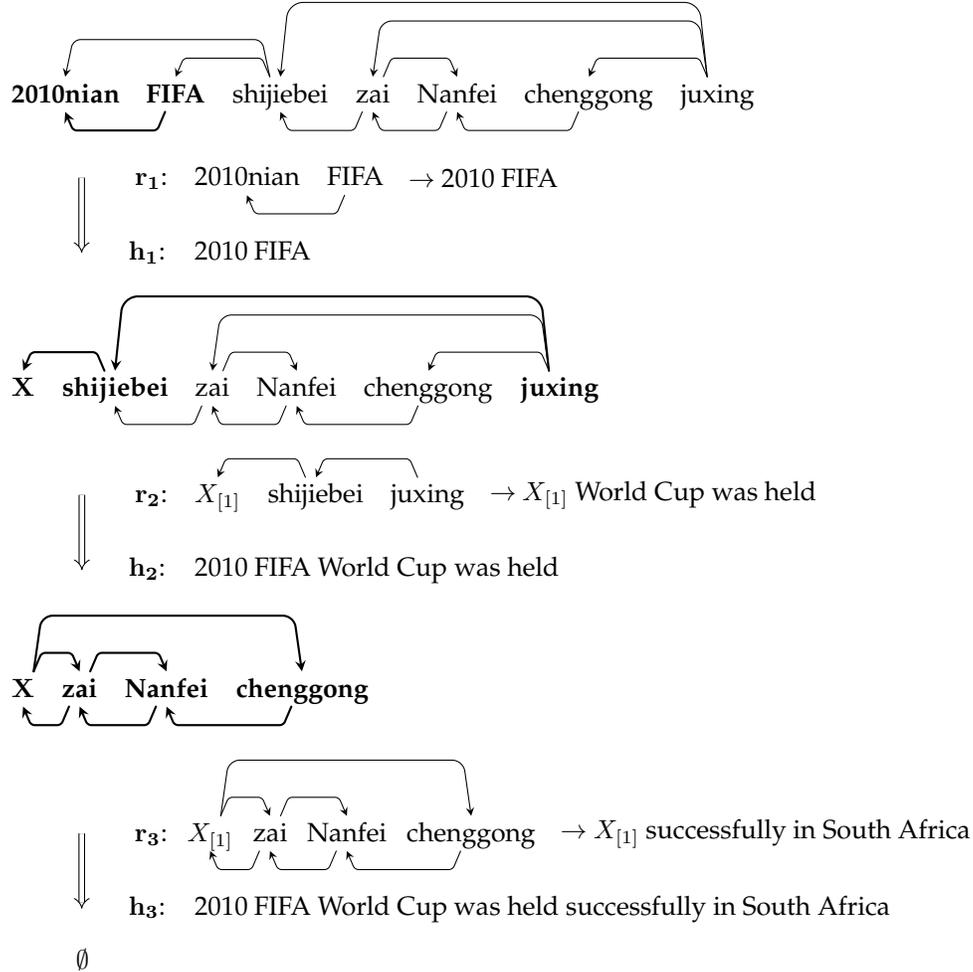

    \centering
    
    \begin{tabular}{l}
        
        \begin{dependency}[hide label,edge unit distance=1.5ex]
            \begin{deptext}[column sep=0.2cm]
            \bf 2010nian \& \bf FIFA \& shijiebei \& zai \& Nanfei \& chenggong \& juxing\\
            \end{deptext}
            \depedge {3}{1}{}
            \depedge {3}{2}{}
            \depedge {7}{3}{}
            \depedge {7}{4}{}
            \depedge {7}{6}{}
            \depedge {4}{5}{}
            \depedge [edge below, thick] {2}{1}{}
            \depedge [edge below] {4}{3}{}
            \depedge [edge below] {5}{4}{}
            \depedge [edge below] {6}{5}{}
            %
            % \node [left = 0cm and 0cm of \matrixref] {\it Input:};
        \end{dependency}
        \\
        \hspace{1cm}
        \begin{dependency}[hide label,edge unit distance=1.5ex]
            \begin{deptext}[column sep=0.2cm]
            $\mathbf{r_1}$: \& 2010nian \& FIFA \\
            \end{deptext}
            \depedge [edge below] {3}{2}{}
            
            \node (t) [right =0 and 0 of \matrixref] {$\to$ 2010 FIFA};
            % \draw [arrow,dashed,thick] (\matrixref) -- (t);
            \node (h) [below =of \matrixref.west,anchor=west] (h) {$\mathbf{h_1}$:\hspace{0.3cm} 2010 FIFA};
            
            \draw [-implies,double equal sign distance] ([xshift=-0.5cm]\matrixref.west) -- ([xshift=-0.5cm]h.west);
        \end{dependency}
        \\
        \begin{dependency}[hide label,edge unit distance=1.5ex]
            \begin{deptext}[column sep=0.2cm]
            $\mathbf{X}$ \& \bf shijiebei \& zai \& Nanfei \& chenggong \& \bf juxing\\
            \end{deptext}
            \depedge [thick]{2}{1}{}
            \depedge [thick] {6}{2}{}
            \depedge {6}{3}{}
            \depedge {6}{5}{}
            \depedge {3}{4}{}
            \depedge [edge below] {3}{2}{}
            \depedge [edge below] {4}{3}{}
            \depedge [edge below] {5}{4}{}
            %
            % \node [left = 0cm and 0cm of \matrixref] {\it Input:};
        \end{dependency}
        \\
        \hspace{1cm}
        \begin{dependency}[hide label,edge unit distance=1.5ex]
            \begin{deptext}[column sep=0.2cm]
            $\mathbf{r_2}$: \& $X_{[1]}$ \& shijiebei \& juxing\\
            \end{deptext}
            \depedge {3}{2}{}
            \depedge {4}{3}{}
            
            \node (t) [right =0 and 0 of \matrixref] {$\to$ $X_{[1]}$ World Cup was held};
            % \draw [arrow,dashed,thick] (\matrixref) -- (t);
            \node (h) [below =of \matrixref.west,anchor=west] (h) {$\mathbf{h_2}$:\hspace{0.3cm} 2010 FIFA World Cup was held};
            
            \draw [-implies,double equal sign distance] ([xshift=-0.5cm]\matrixref.west) -- ([xshift=-0.5cm]h.west);
        \end{dependency}
        \\
        \begin{dependency}[hide label,edge unit distance=1.5ex]
            \begin{deptext}[column sep=0.2cm]
            $\mathbf{X}$ \& \bf zai \& \bf Nanfei \& \bf chenggong\\
            \end{deptext}
            \depedge [thick] {2}{3}{}
            \depedge [edge below,thick] {3}{2}{}
            \depedge [edge below, thick] {4}{3}{}
            \depedge [thick]{1}{2}{}
            \depedge [thick,edge below]{2}{1}{}
            \depedge [thick]{1}{4}{}
            
            % \node [left = 0cm and 0cm of \matrixref] {\it Input:};
        \end{dependency}
        \\
        \hspace{1cm}
        \begin{dependency}[hide label,edge unit distance=1.5ex]
            \begin{deptext}[column sep=0.1cm]
            $\mathbf{r_3}$: \& $X_{[1]}$ \& zai \& Nanfei \& chenggong\\
            \end{deptext}
            \depedge {3}{4}{}
            \depedge [edge below] {4}{3}{}
            \depedge [edge below] {5}{4}{}
            \depedge {2}{3}{}
            \depedge [edge below]{3}{2}{}
            \depedge {2}{5}{}
            
            \node (t) [right =0 and 0 of \matrixref] {$\to$ $X_{[1]}$ successfully in South Africa};
            % \draw [arrow,dashed,thick] (\matrixref) -- (t);
            \node (h) [below =of \matrixref.west,anchor=west] (h) {$\mathbf{h_3}$:\hspace{0.3cm} 2010 FIFA World Cup was held successfully in South Africa};
            
            \draw [-implies,double equal sign distance] ([xshift=-0.5cm]\matrixref.west) -- ([xshift=-0.5cm]h.west);
        \end{dependency}
        \\
        \hspace{0.85cm}
        \begin{dependency}
            \node(a){$\emptyset$};
        \end{dependency}
    \end{tabular}
    
    \caption {An example derivation in our SNRG-based model which parses a DBG and generates an English string in a bottom-up manner. $r_i$ are rules while $h_i$ are hypotheses. Indexes on non-terminals of rules indicate mappings. }
    \label{fig:snrg:derivation}
\end{figure}

%% file: input/alg-snrgdecoderbeam.tex
\begin{algorithm}[t]
\caption{Beam-search decoder for \snrg}
\label{alg:snrgdecode:beam}

\begin{algorithmic}[1]
    \ForAll{rules $X\to\langle\gamma,\alpha\rangle$}
        \State add $(X\to\gamma)$ to {\it Axiom}
    \EndFor
    \For{$l=1$ to $|s|$}
        % \ForAll{$i,j:j-i=l$}
            \If{$l\le L_{\mathrm{max}}$}
                \ForAll{items $[X, \tilde{s}]:w$ s.t. $|\tilde{s}|=l$ inferable from {\it Axiom} and {\it chart}}
                    \State add $[X, \tilde{s}]$ to $B[X,l]$
                    \State recombine and prune if applicable
                \EndFor
            \EndIf
            % \If{$i=0$}
                \ForAll{items $[S, \tilde{s}]:w$ s.t. $|\tilde{s}|=l$ inferable from {\it Axiom} and {\it chart}}
                    \State add $[S, \tilde{s}]$ to $B[S,l]$
                    \State recombine and prune if applicable
                \EndFor
            % \EndIf
        % \EndFor
        \ForAll {items $[x,\tilde{s}]\in B[x,l]$ s.t. $x\in\{X,S\}$}
            \State add $[x, \tilde{s}]$ to $chart[x,\tilde{s}]$
        \EndFor
    \EndFor
    % \Return $\argmax B[S,|s|]$
\end{algorithmic}

\end{algorithm}

%% file: input/alg-snrgdecoderchart.tex
\begin{algorithm}[t]
\caption{Chart decoder \citep{hpb:chiang:2007} for \snrg.}
\label{alg:snrgdecode:chart}

\begin{algorithmic}[1]
    \ForAll{rules $X\to\langle\gamma,\alpha\rangle$}
        \State add $(X\to\gamma)$ to {\it Axiom}
    \EndFor
    \For{$l=1$ to $|s|$}
        \ForAll{$i,j:j-i=l$}
            \If{$l\le g_{\mathrm{max}}$}
                \ForAll{items $[X, i, j]:w$ inferable from {\it Axiom} and {\it chart}}
                    \State add $[X, i,j]$ to $chart[X,i,j]$
                \EndFor
            \EndIf
            \If{$i=0$}
                \ForAll{items $[S, i,j]:w$ inferable from {\it Axiom} and {\it chart}}
                    \State add $[S, i,j]$ to $chart[S,i,j]$
                \EndFor
            \EndIf
        \EndFor
    \EndFor
    % \Return $\argmax chart[S,0,|s|]$
\end{algorithmic}

\end{algorithm}

%% file: input/tab-data.tex
\begin{table}
\centering
\caption{ZH--EN and DE--EN corpora. Word counts are averaged across all references.}
\label{tab:data}
\begin{tabular}{clrrrr}
\hline
\noalign{\smallskip}
&	& \#Sentences & \#Words (ZH) & \#Words (EN) \\ 
\noalign{\smallskip}
\hline
\noalign{\smallskip}
\multirow{4}{*}{ZH--EN} & Train & 1.5M+ & 38M+ & $\sim$45M  \\
& MT02 & 878 &	 22,655 & 26,905  \\
& MT04 & 1,597 &	 43,719 & 52,705  \\
& MT05 & 1,082 &	 29,880 & 35,326  \\
\hline
\noalign{\smallskip}
\multirow{4}{*}{DE--EN} & Train & 2M+ &	52M+ & 55M+  \\
& WMT11 & 3,003 &	 72,661 & 74,753  \\
& WMT12 & 3,003 &	 72,603 & 72,988  \\
& WMT13 & 3,000 &	 63,412 & 64,810  \\ \hline
\end{tabular}
\end{table}

%% file: input/tab-mainscore.tex
\begin{table}
\centering
\caption {BLEU scores of all systems on all test sets. $*$ means a system is significantly better than PBMT at $p\le0.01$. $+$ means a system is significantly better than HPBMT at $p\le0.01$.}\label{tab:mainscore}
\begin{tabular}{lllllll}
\hline
\noalign{\smallskip}
\multirow{2}{*}{System} & \multirow{2}{*}{Graph} & \multirow{2}{*}{\#Rules} & \multicolumn{2}{c}{ZH--EN} & \multicolumn{2}{c}{DE--EN} \\
& & & MT04 & MT05 & WMT12 & WMT13 \\
\noalign{\smallskip}
\hline
\noalign{\medskip}

\multicolumn{7}{c}{Segmentation-based models}\\
\noalign{\medskip}

PBMT & & 69M/107M
& 33.2 & 31.8 & 19.5 & 21.9 \\
TBMT & & 122M/151M
& 33.8$^{*}$ & 31.4 & 19.6 & 22.2$^{*}$ \\
SDTU & & 224M/352M
& 34.7$^{*}$ & 32.6$^{*}$ & 19.7$^{*}$ & 22.4$^{*}$ \\

% \cline{4-7}
\noalign{\smallskip}

\multirow{3}{*}{\seggbmt} 
& Tree & 42M/73M
& 32.1 & 30.9 & 18.1 & 20.4 \\
& DSG & 62M/93M
& 34.2$^{*}$ & 32.7$^{*}$ & 20.0$^{*}$ & 22.5$^{*}$ \\
& DBG & 99M/153M
& 34.7$^{*}$ & 32.4$^{*}$ & 20.1$^{*}$ & 22.9$^{*}$ \\
\hline
\noalign{\medskip}

\multicolumn{7}{c}{Synchronous grammar-based models}\\
\noalign{\medskip}

HPBMT & & 388M/684M
& 36.5$^{*}$ & 34.3$^{*}$ & 20.5$^{*}$ & 23.0$^{*}$ \\
Dep2Str & & 84M/92M
& 36.6$^{*}$ & 34.9$^{*+}$ & 20.4$^{*}$ & 22.7$^{*}$ \\
SERG & DEG & 131M/98M
& 36.7$^{*+}$ & 34.8$^{*+}$ & 20.2$^{*}$ & 22.8$^{*}$ \\

% \cline{4-7}
\noalign{\smallskip}

\multirow{3}{*}{\snrg\textsubscript{b}}
& Tree & 188M/380M
& 34.8$^{*}$ & 32.8$^{*}$ & 20.0$^{*}$ & 22.7$^{*}$ \\
& DSG & 423M/684M
& 34.6$^{*}$ & 33.1$^{*}$ & 20.3$^{*}$ & 22.9$^{*}$ \\
& DBG & 1,141M/1,932M
& 35.4$^{*}$ & 33.1$^{*}$ & 19.1 & 21.5 \\

% \cline{4-7}
\noalign{\smallskip}

\multirow{3}{*}{\snrg\textsubscript{c}}
& Tree & 76M/160M
& 36.9$^{*+}$ & 34.2$^{*}$ & 20.6$^{*}$ & 23.0$^{*}$ \\
& DSG & 157M/241M
& 36.9$^{*+}$ & 34.5$^{*+}$ & 20.7$^{*+}$ & 23.4$^{*+}$ \\
& DBG & 415M/740M
& 37.0$^{*+}$ & 34.9$^{*+}$ & 20.6$^{*}$ & 23.2$^{*+}$ \\
\hline
\end{tabular}
\end{table}

%% file: input/eg-pb.tex
\begin{figure}
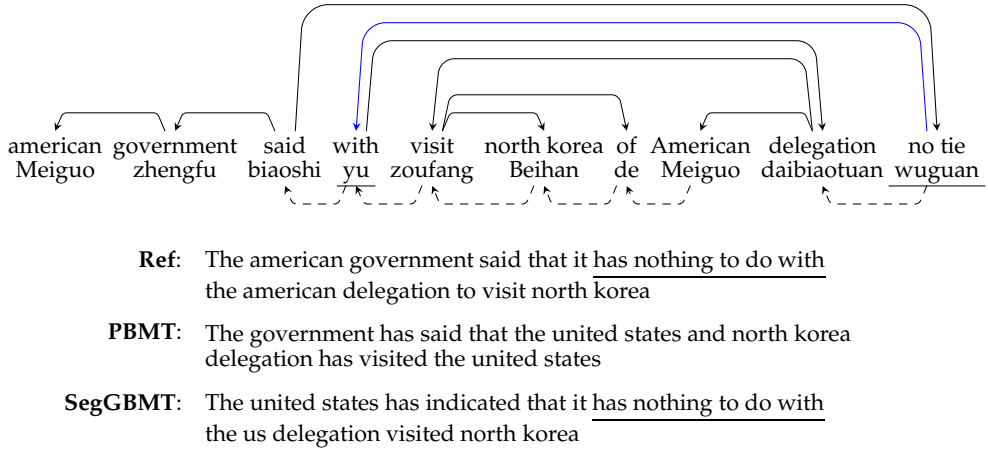

    \centering
    \small
    \scalebox{0.97}{
    \begin{dependency}[hide label]
         \begin{deptext}[column sep=0]
             american \& government \&  said \& with \& visit  \& north korea \& of \& American \& delegation \& no tie \\
             Meiguo \& zhengfu \& biaoshi \& yu \& zoufang \& Beihan \& de \& Meiguo \& daibiaotuan \& wuguan \\
        \end{deptext}
        \depedge {2}{1}{}
        \depedge {3}{2}{}
        \depedge [blue] {10}{4}{}
        \depedge {9}{5}{}
        \depedge {5}{6}{}
        \depedge {5}{7}{}
        \depedge {9}{8}{}
        \depedge {4}{9}{}
        \depedge {3}{10}{}
        
        \depedge [edge below, dashed] {4}{3}{}
        \depedge [edge below, dashed] {5}{4}{}
        \depedge [edge below, dashed] {6}{5}{}
        \depedge [edge below, dashed] {7}{6}{}
        \depedge [edge below, dashed] {8}{7}{}
        \depedge [edge below, dashed] {10}{9}{}
        
        \draw (\wordref{2}{4}.south west) -- (\wordref{2}{4}.south east);
        \draw (\wordref{2}{10}.south west) -- (\wordref{2}{10}.south east);
        
        \node (rlabel) [below= 1cm and 0 of \wordref{2}{2}.south,anchor=east, xshift=0.2cm] {{\bf Ref}:};
        \node (plabel) [below= 1cm and 0 of rlabel.east,anchor=east] {{\bf PBMT}:};
        \node (glabel) [below= 1cm and 0 of plabel.east,anchor=east] {{\bf SegGBMT}:};
        
        \node [right =0 and 0.1 of rlabel.north east, anchor=north west,align=left] {The american government said that it \underline{has nothing to do with}\\ the american delegation to visit north korea};
        \node [right =0 and 0.1 of plabel.north east, anchor=north west,align=left] {The government has said that the united states and north korea\\ delegation has visited the united states};
        \node [right =0 and 0.1 of glabel.north east, anchor=north west, align=left] {The united states has indicated that it \underline{has nothing to do with}\\ the us delegation visited north korea};
        
    \end{dependency}
    }
    \caption{Examples of translations from SegGBMT and PBMT. SegGBMT successfully translated a Chinese collocation (underlined) into a target phrase. PBMT failed to capture this generalization because it only uses continuous phrases.}
    \label{fig:gbmt:eg-pb}
\end{figure}

%% file: input/eg-treelet.tex
\begin{figure}
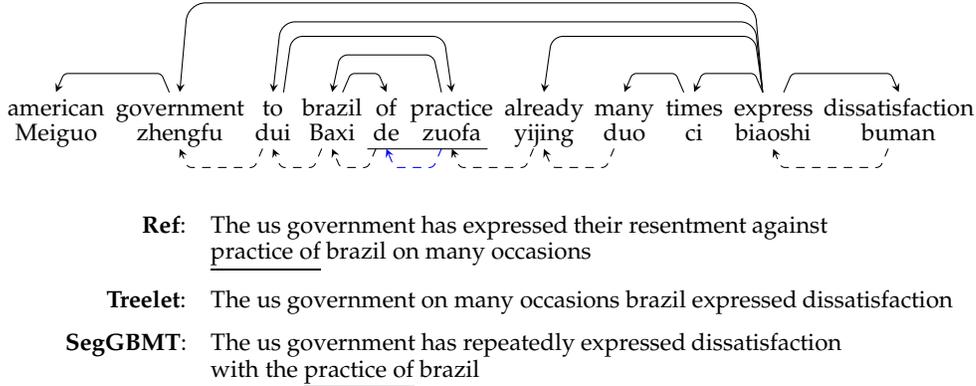

    \centering
    \small
    
    \begin{dependency}[hide label]
         \begin{deptext}[column sep=0]
             american \& government \&  to \& brazil \& of  \& practice \& already \& many \& times \& express \& dissatisfaction \\
             Meiguo \& zhengfu \& dui \& Baxi \& de \& zuofa \& yijing \& duo \& ci \& biaoshi \& buman \\
        \end{deptext}
        \depedge {2}{1}{}
        \depedge [edge unit distance=1.0ex] {10}{2}{}
        \depedge [edge unit distance=0.9ex] {10}{3}{}
        \depedge {6}{4}{}
        \depedge {4}{5}{}
        \depedge {3}{6}{}
        \depedge {10}{7}{}
        \depedge {9}{8}{}
        \depedge {10}{9}{}
        \depedge  {10}{11}{}
        
        \depedge [edge below, dashed] {3}{2}{}
        \depedge [edge below, dashed] {4}{3}{}
        \depedge [edge below, dashed] {5}{4}{}
        \depedge [edge below, dashed, blue] {6}{5}{}
        \depedge [edge below, dashed] {7}{6}{}
        \depedge [edge below, dashed] {8}{7}{}
        \depedge [edge below, dashed] {11}{10}{}
        
        \draw (\wordref{2}{5}.south west) -- (\wordref{2}{6}.south east);

        \node (rlabel) [below= 1cm and 0 of \wordref{2}{2}.south,anchor=east, xshift=0.2cm] {{\bf Ref}:};
        \node (tlabel) [below= 1cm and 0 of rlabel.east,anchor=east] {{\bf Treelet}:};
        \node (glabel) [below= 0.6cm and 0 of tlabel.east,anchor=east] {{\bf SegGBMT}:};
        
        \node [right =0 and 0.1 of rlabel.north east, anchor=north west,align=left] {The us government has expressed their resentment against\\ \underline{practice of} brazil on many occasions};
        \node [right =0 and 0.1 of tlabel.north east, anchor=north west,align=left] {The us government on many occasions brazil expressed dissatisfaction};
        \node [right =0 and 0.1 of glabel.north east, anchor=north west, align=left] {The us government has repeatedly expressed dissatisfaction\\ with the \underline{practice of} brazil};
        
    \end{dependency}
    
    \caption{Examples of translations from SegGBMT and TBMT. By using bigram links, SegGBMT successfully translated the underlined continuous phrase which is not connected in the dependency tree.}
    \label{fig:gbmt:eg-treelet}
\end{figure}

%% file: input/tab-edgelabel.tex
\begin{table}
\centering
\caption{Evaluation results when edges are labeled by their relation types: either dependency or sequential. $*$ means a system is significantly better than its counterpart at $p\le 0.01$.}\label{tab:edgelabel}

\begin{tabular}{lllllll}
\hline
\noalign{\smallskip}
\multirow{2}{*}{System} & \multirow{2}{*}{Graph}& \multirow{2}{*}{\#Rules} & \multicolumn{2}{c}{ZH--EN} & \multicolumn{2}{c}{DE--EN} \\
& & & MT04 & MT05 & WMT12 & WMT13 \\
\noalign{\smallskip}
\hline
\noalign{\smallskip}
SegGBMT & DBG & 99.2M/153.4M
& 34.7 & 32.4 & 20.1 & 22.9 \\
\hspace{5mm}+EdgeLabel & & 99.7M/153.8M
& 34.7 & 32.7$^*$ & 20.1 & 22.9 \\

\noalign{\smallskip}

\snrg\textsubscript{c} & DSG & 157M/241M
& 36.9 & 34.5 & 20.7 & 23.4 \\
\hspace{5mm}+EdgeLabel & & 160M/243M
& 36.7 & 34.7 & 20.6 & 23.3 \\

% \noalign{\smallskip}

% \snrg\textsubscript{chart} & DBG & 415M/740M
% & 37.0 & 34.9 & 20.6 & 23.2 \\
% \hspace{5mm}+EdgeLabel & &
% &  &  &  &  \\
\hline
\end{tabular}
\end{table}

%% file: input/tab-posnt.tex
\begin{table}
\centering
\caption {Evaluation results when linguistic non-terminals are used (denoted as +POS). $*$ means a system is significantly better than its counterpart with or without POS tags at $p\le 0.01$. MGS means minimum size of gaps which can be represented by non-terminals during training.}
\label{tab:posnt}

\begin{tabular}{lllllll}
\hline
\noalign{\smallskip}
\multirow{2}{*}{System} & \multirow{2}{*}{Graph} & \multirow{2}{*}{\#Rules} & \multicolumn{2}{c}{ZH--EN} & \multicolumn{2}{c}{DE--EN} \\
& & & MT04 & MT05 & WMT12 & WMT13 \\
\noalign{\smallskip}
\hline

\noalign{\medskip}

\multicolumn{7}{c}{MGS=2} \\

\noalign{\medskip}
SERG & DEG & 131M/98M
& 36.7 & 34.8 & 20.2 & 22.8 \\
\hspace{5mm}+POS & & 153M/180M
& 36.8 & 34.8  & 20.6$^{*}$ & 23.3$^{*}$ \\

\noalign{\smallskip}

% \snrg\textsubscript{chart} & Tree & 76M/160M
% & 36.9 & 34.2 & 20.6 & 23.0 \\
% \hspace{5mm}+POS & & 79M/172M
% & 36.8 & 34.5 & 20.5 & 23.2$^*$ \\

% \noalign{\smallskip}

\snrg\textsubscript{c} & DSG & 157M/241M
& 36.9 & 34.5 & 20.7 & 23.4 \\
\hspace{5mm}+POS & & 185M/276M
& 36.8 & 34.6  & 20.7 & 23.4 \\

% \noalign{\smallskip}

% \snrg\textsubscript{chart} & DBG & 415M/740M
% & 37.0 & 34.9 & 20.6 & 23.2 \\
% \hspace{5mm}+POS & & 
% &  &   &  &  \\

\hline
\noalign{\medskip}

\multicolumn{7}{c}{MGS=1} \\

\noalign{\medskip}

SERG & DEG & 265M/203M
& 37.0 & 34.9 & 20.1 & 22.8 \\
\hspace{5mm}+POS & & 318M/385M
& 37.7$^*$ & 35.8$^*$  & 20.6$^*$ & 23.2$^*$ \\

\noalign{\smallskip}

% \snrg\textsubscript{chart} & Tree & 167M/333M
% & 37.1$^*$ & 35.3$^*$ & 20.6$^*$ & 23.2 \\
% \hspace{5mm}+POS & & 185M/363M
% & 33.7 & 32.0 & 20.4 & 23.2 \\

% \noalign{\smallskip}

\snrg\textsubscript{c} & DSG & 312M/480M
& 37.2 & 34.7 & 20.7 & 23.6$^*$ \\
\hspace{5mm}+POS & & 382M/563M
& 37.7$^*$ & 35.8$^*$  & 20.7 & 23.4 \\

\hline
\end{tabular}
\end{table}